\pdfoutput=1
\documentclass[twoside,11pt]{article}

\usepackage{amsmath}
\usepackage{amsthm}

\usepackage{style/paperstyle}

\usepackage{mathtools}
\usepackage{bm}
\usepackage{booktabs}
\usepackage{xcolor}
\usepackage{lastpage}

\newtheorem{assumption}[theorem]{Assumption}

\makeatletter
\@addtoreset{theorem}{section}
\makeatother

\makeatletter
\renewenvironment{proof}[1][\proofname]{%
  \par\medskip\noindent{\bfseries #1\@addpunct{.}}\ \ignorespaces
}{%
  \hfill\BlackBox\\[2mm]%
}
\makeatother
\providecommand{\proofname}{Proof}
\renewcommand{\proofname}{Proof}

\providecommand{\qed}{\hfill\BlackBox\\[2mm]}

\newcommand{\R}{\mathbb{R}}
\newcommand{\E}{\mathbb{E}}
\newcommand{\Fnorm}[1]{\lVert #1 \rVert_{F}}
\newcommand{\opnorm}[1]{\lVert #1 \rVert_{\mathrm{op}}}
\renewcommand{\Pr}{\mathcal{P}}
\DeclareMathOperator{\rank}{rank}
\DeclareMathOperator*{\argmin}{arg\,min}

\titlepagestyle

\ShortHeadings{Dimension-Free Buffer Bounds for Error Feedback}{Moayedikia}
\firstpageno{1}

\begin{document}
\sloppy
\setlength{\emergencystretch}{3em}

\title{Dimension-Free Buffer Bounds for\\ Compressed-Momentum Error Feedback}

\author{\name Alireza Moayedikia\thanks{Alireza Moayedikia is with the
Department of Business Technology and Entrepreneurship, Swinburne
University of Technology, John Street, Hawthorn, VIC 3122, Australia
(e-mail: amoayedikia@swin.edu.au). (Corresponding author:
Alireza Moayedikia.)}\\[4pt]
\normalsize\mdseries\upshape Department of Business Technology and Entrepreneurship\\
\normalsize Swinburne University of Technology}

\maketitle

\begin{abstract}%
Analyses of error feedback bound the memory buffer through a compressor
contraction constant --- $r/p$ for the best rank-$r$ approximation of a
$p$-dimensional spectrum --- which survives into the rate. We show it need
not appear at all. When the \emph{compressed} momentum is re-injected into the
pre-compression accumulator, the previous compressed momentum $\beta_1M_{t-1}$
is an admissible competitor in the projection, and the buffer obeys
$\Fnorm{E_t}\le(1-\beta_1)\Fnorm{G_t}+\gamma\Fnorm{E_{t-1}}$, free of the
dimensions $m,n$ and of the rank $r$. The argument uses only that the compressor
is a metric projection onto a closed cone, so it covers top-$k$ selection as
well as truncated \textsc{svd}, holds for every selection from the projection
set with no condition on singular-value crossings, and in its
trajectory-tracking form assumes nothing about the gradients. In the two-stage
template of the standard analyses it replaces a Stage-2 constant of order
$p^2/r^2$ by one of order $1$, against $\Theta(p/r)$ for the sharpest
contraction-based bounds. Transferring the full rate costs a leakage
term, which we bound by the step-weighted gradient mass
$\sum_{u\le t}\eta_u\Fnorm{G_u}$ and show tight up to a factor $\sqrt2$; the
price is an additive $O(\log^2T)$ floor, not a slower rate. On \textsc{bert}
fine-tuning at rank ratio $32/768$ the feedback decay costs under $0.008$
validation $F_1$.
\end{abstract}


\section{Introduction}\label{sec:intro}

An optimizer that compresses its own state must decide what to do with what the
compression discards. Error feedback keeps it in a memory buffer and adds it
back at the next step, a device that goes back to 1-bit
\textsc{sgd}~\citep{seide2014} and whose first convergence analyses are due to
\citet{stich2018} and \citet{karimireddy2019}. Both analyses, and every descendant we are aware
of, control the buffer the same way: they invoke the defining property of a
contractive compressor, $\E\lVert\mathcal C(x)-x\rVert^2\le(1-\alpha)\lVert
x\rVert^2$, and carry $\alpha$ into the rate.

For a low-rank compressor this is expensive, and expensive in a way that sits
oddly with why such compressors are used at all. Low-rank schemes are motivated
by spectral structure that the dynamics themselves produce --- the gradient
becomes low rank during training~\citep[Lemma~3.3]{zhao2024}, and its energy
concentrates in a small subspace~\citep{gurari2018,sagun2018} --- yet the
contraction constant they inherit is quantified over all inputs and knows
nothing of that structure. The best rank-$r$ approximation of a $p$-dimensional
spectrum belongs to the biased-compressor class of \citet{richtarik2021} with $\alpha=r/p$, and the constant is attained
(Corollary~\ref{cor:class}). The classical buffer lemmas
then contribute a factor of order $p^2/r^2$, and the sharpest current
contraction-based leading terms a factor of order $p/r$; at the width and rank
of a \textsc{bert} attention matrix, $p=768$ and $r=32$, those are $2208$ and
$24$ respectively. Neither number says anything about why such optimizers work
in practice, and the second grows without bound as models widen.

The constant need not enter at all, and this paper isolates the structural
property of the recursion that keeps it out. The recursion in question is not
hypothetical: it is the momentum--error-feedback core of the \textsc{umtam}
optimizer of \citet{moayedikia2026}, extracted here for analysis on its own.
Write the pre-compression accumulator as
$B_t=\beta_1M_{t-1}+(1-\beta_1)G_t+\gamma E_{t-1}$, compress it to
$M_t=\Pr_r(B_t)$, and keep the residual $E_t=B_t-M_t$. Because $\Fnorm{E_t}$ is
by definition the \emph{minimum} distance from $B_t$ to the set of rank-$r$
matrices, any rank-$r$ matrix furnishes an upper bound --- and the algorithm
supplies one for nothing. The matrix $\beta_1M_{t-1}$ has rank at most $r$ by
construction, so
\begin{equation}\label{eq:competitor}
  \Fnorm{E_t}\;\le\;\Fnorm{B_t-\beta_1M_{t-1}}
  \;=\;\Fnorm{(1-\beta_1)G_t+\gamma E_{t-1}} ,
\end{equation}
and the momentum term has vanished from the residual without any contraction
being invoked. Everything in \S\ref{sec:bound} and \S\ref{sec:conseq} follows
from \eqref{eq:competitor}.

The property is structural rather than incidental, and \S\ref{ssec:necessity}
says exactly which schemes have it. Residual schemes carry the discarded part
forward but never the compressed object, so no competitor is available.
Difference schemes such as EF21~\citep{richtarik2021} and its momentum variant
\citep{fatkhullin2023} carry a state forward but subtract it, and their
compression residual is an identity rather than a minimisation, so there is
nothing to choose. Gradient-space low-rank
methods~\citep{vogels2019,zhao2024,chen2025} compress before the momentum is
formed. What is required is that the compressed momentum re-enter additively,
before the next truncation.

The argument also uses less about the compressor than one might expect. It needs
only that $\Pr$ is a metric projection onto a set closed under nonnegative
scaling. Rank-$\le r$ matrices form such a cone; so do $k$-sparse vectors, and
top-$k$ selection is the corresponding projection. The bound is therefore not a
fact about the singular value decomposition, and top-$k$ --- the canonical
compressor of this literature, for which the classical constant is $d/k$ ---
obeys it just as well (Remark~\ref{rem:topk}). What is stuck at $d/k$ is not the
compressor but the algorithms in which it is customarily used.

What follows from \eqref{eq:competitor} divides into a bound on the buffer, its
consequences for rates, and a set of measurements. Theorem~\ref{thm:Ebound}
bounds the error buffer by a quantity involving neither the dimensions nor the
rank. It holds for any
cone-projection compressor and for every selection from the projection set, so
that no hypothesis on singular-value crossings along the trajectory is required
(Remark~\ref{rem:selection}), and in its trajectory-tracking form it uses no
assumption on the gradients at all (Remark~\ref{rem:bgfree}). Feeding this into
the standard template, Corollary~\ref{cor:free} closes Stage~2 with a constant
of order $1$ where the classical buffer lemmas carry $p^2/r^2$, and
Corollary~\ref{cor:shift} shows that the step-size shift requirement becomes
dimension-free where \citet{stich2018} need $a=\Theta(d/k)$.
Stage~1 is a different matter and we do not claim it for free:
Proposition~\ref{prop:pi} identifies the perturbed-iterate deviation exactly and
finds in it an accumulated-leakage term absent from both references, and
Theorem~\ref{thm:floor} shows that when this term is bounded rather than
shrinking with the step size, the price is an additive floor rather than a
slower rate. The empirical side supplies three things. The distortion ratio on
\textsc{bert} fine-tuning converges to $0.26$ against a worst case of $0.979$, a
factor of $22$ in $\alpha$ (\S\ref{ssec:transformer}); the $\gamma$ crossover
predicted by Theorem~\ref{thm:bounded} is visible on real gradients and
replicates when the compression drives training rather than merely observing it
(\S\ref{ssec:quality}); and no cost from the feedback decay appears on
\textsc{bert} at the rank ratio a transformer uses, for a reason that
instrumentation makes plain (\S\ref{ssec:quality}).

Two limitations shape what can be concluded, and both point at the natural
continuations of the work. The first is the size of the floor rather than its
existence. Theorem~\ref{thm:leak} bounds the accumulated leakage by the
step-weighted gradient mass $\Gamma_t=\sum_{u\le t}\eta_u\Fnorm{G_u}$, which
under the schedule both reference analyses use grows like $\log t$, so the floor
is $O(\log^2T)$ rather than a constant; and
Proposition~\ref{prop:lower} shows this is not an artefact of the argument, by
exhibiting a two-dimensional top-$k$ instance with a constant gradient in which
the deviation genuinely diverges at that rate. The floor is a true constant
exactly when $\Gamma_\infty$ converges, which is a condition on how fast the
gradients decay along the trajectory and not on the compressor
(Corollary~\ref{cor:Dfinite}); what is still missing is the sharper argument
that would explain why simulation shows the leakage growing some five hundred
times more slowly than the bound when the discarded tails are not aligned.
The second limitation concerns the gap between the floor as a proved consequence
and the floor as a measured effect. Instrumenting the network shows the deviation
to be roughly half the fine-tuning displacement but only a thousandth of
the pretrained weights, and it is the second ratio that validation quality
responds to; the floor is therefore invisible at this scale of displacement
without the theorem being in any doubt. That reconciliation depends on fine-tuning
moving a short distance from a good initialisation, so it says nothing about
training from scratch, where the displacement is a large fraction of the weights
and the floor should be expected to bite. Establishing whether it does is left to
future work.

\section{Setting}\label{sec:setting}

\subsection{The recursion}\label{ssec:recursion}

Fix $m,n$ and write $p=\min\{m,n\}$. For $B\in\R^{m\times n}$ with singular value
decomposition $B=\sum_{i=1}^{p}\sigma_i(B)\,u_iv_i^{\top}$ and
$\sigma_1(B)\ge\cdots\ge\sigma_p(B)\ge 0$, let
\begin{equation}\label{eq:proj}
  \Pr_r(B) \;=\; \sum_{i=1}^{r}\sigma_i(B)\,u_iv_i^{\top}
\end{equation}
denote a best rank-$r$ approximation in the Frobenius norm (Eckart--Young). The
map $\Pr_r$ is the metric projection onto the determinantal variety
$\R^{m\times n}_{\le r}$, and it is properly a \emph{set}-valued map: by
Eckart--Young, $\Pr_r(B)$ is the set of all rank-$\le r$ truncated singular
value decompositions of $B$, which is nonempty and compact because
$\R^{m\times n}_{\le r}$ is closed~\citep{olikier2024lowrank}. It is a singleton
if and only if $\sigma_r(B)>\sigma_{r+1}(B)$ \emph{or} $\rank B\le r$; the
second case is not vacuous here, since it is exactly the case $E=0$, and it is
missed by the gap condition alone (when $\rank B\le r$ one has
$\sigma_r(B)=\sigma_{r+1}(B)$, possibly $=0$, yet $\Pr_r(B)=\{B\}$). \citet{absil2012} show that $\Pr_r$ is well defined and smooth on a
neighbourhood characterised by $\Fnorm{B-\bar B}<\sigma_r(\bar B)/2$ for
$\bar B\in\mathcal M_r$, and that it is a second-order retraction
there~\citep{absil2015}; the manifold $\mathcal M_r$ and its geometry are treated
in \citet{vandereycken2013,absil2008}.

Following the convention of the variational-geometry literature on this variety
\citep{olikier2024pgd,olikier2024lowrank}, we do not impose a gap condition. We
fix an arbitrary selection, writing $\Pr_r(B)$ for some element of the
projection set, and Remark~\ref{rem:selection} records that every result below
holds for every selection, so that no hypothesis on singular-value crossings
along the trajectory is needed.

The results of \S\ref{sec:bound} use only two properties of $\Pr_r$, so we
record the abstraction now and specialise afterwards.

\begin{definition}[Cone-projection compressor]\label{def:cone}
Let $S\subseteq\R^{m\times n}$ be a nonempty closed \emph{cone}: $X\in S$ and
$\lambda\ge0$ imply $\lambda X\in S$. A compressor $\Pr_S$ is a
\emph{cone projection} onto $S$ if
$\Pr_S(B)\in\argmin\{\Fnorm{B-X}:X\in S\}$ for every $B$.
\end{definition}

\noindent
Two instances matter. Rank-$\le r$ matrices form a closed cone and $\Pr_r$ of
\eqref{eq:proj} is the corresponding projection (Eckart--Young). The
$k$-sparse vectors form a closed cone and top-$k$ selection is the
corresponding projection --- the canonical compressor of the error-feedback
literature. We write $\Pr_r$ throughout, since the low-rank case is the
algorithm of interest, but every statement in \S\ref{sec:bound} holds for
$\Pr_S$ with $S$ any closed cone; \S\ref{ssec:necessity} returns to this.

We study the recursion generated by an optimizer that maintains a rank-$r$
momentum factor together with an error-feedback buffer --- the training core
of the \textsc{umtam} optimizer, whose momentum update with error feedback is
exactly this recursion \citep[Eqs.~(7)--(9)]{moayedikia2026}. With momentum
coefficient $\beta_1\in[0,1)$, feedback decay $\gamma\in[0,1]$, and stochastic
gradients $G_t$, the state $(M_t,E_t)$ evolves as
\begin{equation}\label{eq:recursion}
  B_t \;=\; \beta_1 M_{t-1} + (1-\beta_1)G_t + \gamma E_{t-1},
  \qquad
  M_t \;=\; \Pr_r(B_t),
  \qquad
  E_t \;=\; B_t - M_t,
\end{equation}
initialised at $M_0=E_0=0$. We call $B_t$ the \emph{pre-compression
accumulator}. Note the order of operations: the error buffer is re-injected
\emph{before} truncation, so $E_t$ is the truncation residual of a quantity that
already contains $E_{t-1}$. This is the feature that drives everything below.

For comparison with the error-feedback literature we also use the
step-size-carrying form, in which the step size enters the buffer when the
gradient does rather than when it is retrieved.

\begin{assumption}[Algorithm under analysis]\label{as:alg}
Parameters $X_t\in\R^{m\times n}$ and step sizes $\eta_t>0$ evolve as
\begin{equation}\label{eq:recursion-eta}
  \tilde B_t=\beta_1\tilde M_{t-1}+\eta_t(1-\beta_1)G_t+\gamma\tilde E_{t-1},
  \quad
  \tilde M_t=\Pr_r(\tilde B_t),
  \quad
  \tilde E_t=\tilde B_t-\tilde M_t,
  \quad
  X_{t+1}=X_t-\tilde M_t,
\end{equation}
initialised at $\tilde M_0=\tilde E_0=0$.
\end{assumption}

\noindent
Here $\tilde M_t,\tilde E_t$ are \emph{displacements}; for constant
$\eta_t\equiv\eta$ one has $\tilde M_t=\eta M_t$, $\tilde E_t=\eta E_t$.
Equation~\eqref{eq:recursion-eta} omits the row/column preconditioner of the
full \textsc{umtam} algorithm~\citep{moayedikia2026}; the analysis is therefore
for the unpreconditioned variant.

\begin{assumption}[Bounded gradients]\label{as:G}
There is $G<\infty$ with $\Fnorm{G_t}\le G$ for all $t$.
\end{assumption}

\noindent
Assumption~\ref{as:G} is standard in the error-feedback literature and is
implied by the gradient clipping such optimizers apply in practice.

\subsection{Notation}\label{ssec:notation}

Table~\ref{tab:notation} collects the symbols used throughout. Two conventions
deserve emphasis because they collide with usage elsewhere in the
error-feedback literature. Our $\gamma$ is the \emph{feedback decay}, not a step
size; in \citet{fatkhullin2023} the symbol $\gamma$ denotes the step size and
$\eta$ denotes what we write as $1-\beta_1$. And $\alpha$ always denotes a
compressor contraction constant, never a momentum parameter.

\begin{table}[h]\centering\footnotesize\setlength{\tabcolsep}{2.5pt}
\begin{tabular}{llll}
\toprule
\multicolumn{2}{l}{\emph{Dimensions and compression}} &
\multicolumn{2}{l}{\emph{Algorithm state}} \\
$m,n$ & matrix dimensions & $B_t$ & pre-compression accumulator \\
$p$ & $\min\{m,n\}$ & $M_t$ & compressed momentum, $\Pr_r(B_t)$ \\
$r$ & truncation rank & $E_t$ & error buffer, $B_t-M_t$ \\
$\Pr_r$ & best rank-$r$ approximation & $\tilde B_t,\tilde M_t,\tilde E_t$
  & same, step size carried inside \\
$\mathcal M_r$ & fixed-rank manifold & $X_t$ & parameter iterate \\
$\sigma_i(\cdot)$ & $i$-th singular value & $\hat X_t$ & virtual iterate,
  $X_t-\Delta_{t-1}$ \\
$q(B)$ & distortion ratio $\Fnorm{E}/\Fnorm{B}$ & $\Delta_t$ & deviation
  $\hat X_t$ from $X_t$, Eq.~\eqref{eq:pi} \\
& & $S_t$ & accumulated leakage $\sum_{s\le t}\tilde E_s$ \\
\midrule
\multicolumn{2}{l}{\emph{Parameters}} &
\multicolumn{2}{l}{\emph{Constants}} \\
$\beta_1$ & momentum coefficient & $\alpha$ & compressor contraction constant \\
$\gamma$ & feedback decay & $\alpha_T$ & restricted constant,
  Def.~\ref{def:restricted} \\
$\eta_t$ & step size & $\delta$ & $\alpha$ for $\Pr_r$, equal to $r/p$ \\
$\rho$ & schedule ratio $\sup_t\eta_{t-1}/\eta_t$ & $\kappa$ &
  $\sqrt{\beta_1^2+\gamma^2}$, Thm.~\ref{thm:bounded} \\
$a$ & step-size shift, $\eta_t=c/(a+t)$ & $D$ & $\sup_t\E\Fnorm{\Delta_t}^2$,
  Thm.~\ref{thm:floor} \\
$T$ & horizon; $w_t,W_T$ averaging weights & $G,L,\mu,\sigma$ & gradient bound,
  smoothness, strong convexity, noise \\
& & $D_T$ & $\sup_{t<T}\E\Fnorm{\Delta_t}^2$, Cor.~\ref{cor:Dlog} \\
& & $\Gamma_t$ & $\sum_{u\le t}\eta_u\Fnorm{G_u}$, Eq.~\eqref{eq:Gamma} \\
\bottomrule
\end{tabular}
\caption{Notation.}\label{tab:notation}
\end{table}

Norms are Frobenius unless subscripted otherwise; $\opnorm{\cdot}$ is the
operator norm and $\langle\cdot,\cdot\rangle_F$ the trace inner product. For the
comparison in \S\ref{ssec:stage2} we follow the source notation of
\citet{stich2018}, writing $d$ for the ambient dimension and $k$ for the
sparsification parameter, so that their $d/k$ plays the role of our $p/r$.

\subsection{Relation to existing schemes}\label{ssec:relation}

The recursion \eqref{eq:recursion-eta} differs from the schemes it resembles in
one respect that turns out to carry the whole of \S\ref{sec:bound}: the object
being compressed is the momentum, and the previous \emph{compressed} momentum is
a summand of the quantity that is compressed next. Table~\ref{tab:schemes}
separates the nearby methods on that basis.

\begin{table}[h]\centering\footnotesize\setlength{\tabcolsep}{4pt}
\begin{tabular}{lccc}
\toprule
scheme & compressed object & error feedback & compressed state re-enters \\
\midrule
\textsc{mem-sgd}~\citep{stich2018}          & gradient        & yes & no \\
\citet{karimireddy2019}                      & update          & yes & no \\
EF21~\citep{richtarik2021}                   & gradient        & reference point & subtracted \\
EF21-Muon~\citep{muon2025}                   & momentum diff.  & reference point & subtracted \\
\citet{paramagnostic2025}                    & momentum diff.  & reference point & subtracted \\
PowerSGD~\citep{vogels2019}                  & gradient        & yes & no \\
GaLore~\citep{zhao2024}                      & gradient        & no  & no \\
LDAdam~\citep{robert2025}                    & gradient, state & yes & no \\
EFCP~\citep{modoranu2024}                    & precond.\ window& yes & no \\
GreedyLore~\citep{chen2025}                  & gradient        & yes & no \\
\eqref{eq:recursion-eta}                    & momentum        & yes & \textbf{yes} \\
\bottomrule
\end{tabular}
\caption{Nearby schemes by what they compress and whether the compressed state
is fed back into the pre-compression quantity. Only the last row has that
property in the additive form the argument of \S\ref{ssec:main} needs; the
\textsc{ef21} family re-injects its state, but subtractively, which is a
different thing. \S\ref{ssec:necessity} shows that this distinction is what
supplies the Eckart--Young competitor.}\label{tab:schemes}
\end{table}

Three of the distinctions deserve comment. \textsc{mem-sgd} and its non-convex
counterpart maintain a memory of what compression discarded and add it to the
next gradient, but the memory is the \emph{residual}, never the compressed
object, so the pre-compression quantity contains no low-rank summand.
EF21~\citep{richtarik2021} does carry a state forward --- a reference point
updated by compressed differences --- and is in that sense the closest
conceptual relative; but the reference point is subtracted rather than added to
form the compressed quantity, so it does not serve as a competitor in the
minimisation of \S\ref{ssec:main}. Gradient-space low-rank
methods~\citep{vogels2019,zhao2024,chen2025} compress $G_t$ before momentum is
formed, so the previous compressed object is not a summand of the current one at
all.

Two further differences are recorded for completeness rather than used.
LDAdam~\citep{robert2025} compresses both gradient and optimizer state, and its
analysis is conducted for vector-valued quantities with the matrix case handled
by assuming a Frobenius contraction; the compressor there is a linear projection
onto a fixed subspace, whereas $\Pr_r$ is a nonlinear projection onto the
determinantal variety. EFCP~\citep{modoranu2024} compresses a window of gradients
before it enters a preconditioner, which is architecturally the closest to the
full algorithm this paper analyses, but the compressed window is not itself fed
back.

\section{Spectral structure of the compression}\label{sec:split}

\subsection{Orthogonal splitting}\label{ssec:splitting}

\begin{lemma}[Orthogonal splitting]\label{lem:split}
Let $S\subseteq\R^{m\times n}$ be a nonempty closed cone, let $\Pr_S$ be a cone
projection (Definition~\ref{def:cone}), and set $M=\Pr_S(B)$, $E=B-M$. Then
\begin{equation}\label{eq:split}
  \langle M,E\rangle_F = 0,
  \qquad
  \Fnorm{B}^2=\Fnorm{M}^2+\Fnorm{E}^2 .
\end{equation}
Consequently the \emph{distortion ratio} $q(B):=\Fnorm{E}/\Fnorm{B}$ (defined for
$B\neq0$) satisfies $q(B)\in[0,1]$, $\Fnorm{E}=q(B)\Fnorm{B}$ and
$\Fnorm{M}=\sqrt{1-q(B)^2}\,\Fnorm{B}$.

If in addition $S=\R^{m\times n}_{\le r}$, so that $\Pr_S=\Pr_r$, then
\begin{equation}\label{eq:qexact}
  \Fnorm{E}^2=\sum_{i>r}\sigma_i(B)^2,
  \qquad
  q(B)^2 \;=\; \frac{\sum_{i>r}\sigma_i(B)^2}{\sum_{i\ge 1}\sigma_i(B)^2},
\end{equation}
i.e.\ $q^2$ \emph{equals} the fraction of spectral energy outside the leading
rank-$r$ subspace, with no inequality incurred.
\end{lemma}

\noindent
Proof in Appendix~\ref{app:proofs}. Lemma~\ref{lem:split} applies verbatim to
$(\tilde B_t,\tilde M_t,\tilde E_t)$, being a statement about a single
compression.

Two remarks on the scope of \eqref{eq:split}, which is used repeatedly below.
First, it is a property of \emph{projection onto a cone}, not of the singular
value decomposition: the argument is that $\lambda M\in S$ for every
$\lambda\ge0$, so $\lambda\mapsto\Fnorm{B-\lambda M}^2$ is minimised over
$[0,\infty)$ at the interior point $\lambda=1$ and its derivative vanishes
there. It therefore holds verbatim for top-$k$ selection; we verify this
numerically in \S\ref{ssec:exact}. Second, and consequently, every result below
that consumes Lemma~\ref{lem:split} only through \eqref{eq:split} ---
Lemma~\ref{lem:onestep}, Theorem~\ref{thm:bounded}, and the identity for
$\Fnorm{\Delta_t}$ at $\gamma=1$ in \S\ref{ssec:transfer} --- holds for an
arbitrary cone-projection compressor, exactly as Theorem~\ref{thm:Ebound} does.
Only the spectral form \eqref{eq:qexact} and Corollary~\ref{cor:class} are
specific to truncated \textsc{svd}.

\subsection{Class membership and tightness}\label{ssec:class}

\begin{corollary}[$\Pr_r$ is a contractive compressor, tightly]\label{cor:class}
For every $B\in\R^{m\times n}$,
$\Fnorm{B-\Pr_r(B)}^2 \le (1-r/p)\Fnorm{B}^2$,
so $\Pr_r$ belongs to the biased-compressor class $\mathcal B(\alpha)$ of
\citet{richtarik2021} with $\alpha=r/p$. The constant cannot be improved uniformly: equality holds whenever
$B$ has $p$ equal nonzero singular values.
\end{corollary}

\begin{proof}
By \eqref{eq:qexact}, $q(B)^2=\sum_{i>r}\sigma_i^2/\sum_i\sigma_i^2$. Since the
$\sigma_i$ are non-increasing, the mean of the largest $r$ squared singular
values is at least the mean of all $p$, so
$\sum_{i\le r}\sigma_i^2\ge\frac{r}{p}\sum_{i\le p}\sigma_i^2$, whence
$q(B)^2\le 1-r/p$. The argument is unaffected if $\rank B<p$, since the
additional zeros only lower the overall mean. If
$\sigma_1=\cdots=\sigma_p=\sigma>0$ then
$q(B)^2=(p-r)\sigma^2/(p\sigma^2)=1-r/p$ exactly, so the constant is attained.
\end{proof}

\subsection{Flat spectra and the universal constant}\label{ssec:wrong}

Corollary~\ref{cor:class} shows that truncated-SVD compression requires no new
compressor theory: it sits inside an existing class, with a constant matching
top-$k$'s $\alpha=k/d$~\citep{stich2018,richtarik2021}. What it also shows is that the universal constant is
attained \emph{only} at a flat spectrum. Membership in $\mathcal B(\alpha)$ is a
condition quantified over all inputs, whereas an optimizer visits only the
inputs its own dynamics generate. At $p=768$, $r=32$ the universal constant
gives $q\le0.979$; the measured value on \textsc{bert} fine-tuning is
$q\approx0.26$ (\S\ref{ssec:transformer}), a discrepancy of a factor of $22$ in
$\alpha$.

\subsection{The restricted constant}\label{ssec:restricted}

\begin{definition}[Restricted contraction constant]\label{def:restricted}
For a trajectory $\{B_t\}_{t\le T}$ generated by \eqref{eq:recursion}, set
$\alpha_T := 1-\sup_{t\le T} q(B_t)^2$.
\end{definition}

\noindent
By construction $\alpha_T\ge r/p$, and any guarantee proved for compressors in
$\mathcal B(\alpha)$ that invokes the contraction only at the visited iterates
holds with $\alpha$ replaced by $\alpha_T$. We record $\alpha_T$ because it is
the quantity one would report to compare against the compressor literature on
its own terms; it is not load-bearing for the results below, which control the
buffer directly and never form the ratio.
One point of care: the class condition is used inside a conditional expectation, so the
substitution is legitimate only if $\alpha_T$ is adapted to the filtration, or
if the supremum is taken over a deterministic set containing the trajectory with
high probability. We do not use the substitution in what follows.

\section{The buffer bound}\label{sec:bound}

\subsection{Main result}\label{ssec:main}

\begin{theorem}[Buffer bound for a cone-projection compressor]\label{thm:Ebound}
Let $\Pr_S$ be a cone projection (Definition~\ref{def:cone}) and let
$\gamma\in[0,1]$. Then for the recursion \eqref{eq:recursion} with $\Pr_S$ in
place of $\Pr_r$, initialised at $M_0=E_0=0$,
\begin{equation}\label{eq:Etraj}
  \Fnorm{E_t}\;\le\;(1-\beta_1)\Fnorm{G_t}+\gamma\Fnorm{E_{t-1}}
  \qquad(t\ge1),
\end{equation}
and hence $\Fnorm{E_t}\le(1-\beta_1)\sum_{j=0}^{t-1}\gamma^{j}
\Fnorm{G_{t-j}}$. No assumption on the gradients is used, and the
statement involves no property of $S$ beyond its being a closed cone --- in
particular neither the dimensions $m,n$ nor the rank $r$ appear.
\end{theorem}

\begin{corollary}[Dimension-free limit under bounded gradients]\label{cor:Ebound}
If in addition Assumption~\ref{as:G} holds and $\gamma<1$, then
$\Fnorm{E_t}\le\frac{(1-\beta_1)G}{1-\gamma}(1-\gamma^{t})
\to\frac{(1-\beta_1)G}{1-\gamma}$, a limit independent of $m$, $n$ and $r$.
\end{corollary}

\begin{proof}[Proof of Theorem~\ref{thm:Ebound} and Corollary~\ref{cor:Ebound}]
By Definition~\ref{def:cone},
$\Fnorm{E_t}=\Fnorm{B_t-\Pr_S(B_t)}=\min\{\Fnorm{B_t-X}:X\in S\}$. The matrix
$X=\beta_1M_{t-1}$ is admissible: $M_{t-1}=\Pr_S(B_{t-1})\in S$ by
construction, and $\beta_1\ge0$, so $\beta_1M_{t-1}\in S$ because $S$ is a cone.
Hence, using \eqref{eq:recursion},
\[
  \Fnorm{E_t}\;\le\;\Fnorm{B_t-\beta_1M_{t-1}}
  \;=\;\Fnorm{(1-\beta_1)G_t+\gamma E_{t-1}}
  \;\le\;(1-\beta_1)\Fnorm{G_t}+\gamma\Fnorm{E_{t-1}},
\]
the last step by the triangle inequality; this is \eqref{eq:Etraj}, and no
property of $G_t$ has been used. Unrolling from $\Fnorm{E_0}=0$ gives
$\Fnorm{E_t}\le(1-\beta_1)\sum_{j=0}^{t-1}\gamma^{j}\Fnorm{G_{t-j}}$. For the
corollary, substituting $\Fnorm{G_t}\le G$ (Assumption~\ref{as:G}) and summing
the geometric series gives
$\Fnorm{E_t}\le(1-\beta_1)G(1-\gamma^t)/(1-\gamma)$, and $\gamma<1$ yields the
limit. Neither bound involves $m$, $n$, $r$, or any property of $S$ beyond its
being a closed cone.
\end{proof}

\noindent
Three lines, and the whole content is in the choice of competitor.
$\Fnorm{E_t}$ is by definition the \emph{minimum} distance from $B_t$ to $S$;
the matrix $\beta_1M_{t-1}$ lies in $S$, because $M_{t-1}\in S$ by construction
and $S$ is a cone; so using it as a competitor removes the momentum term from
the residual entirely. Closure under positive scaling is the only structure
required, which is why the statement is not specific to truncated \textsc{svd}.

\begin{remark}[Well-posedness of the projection is not needed]\label{rem:selection}
Theorem~\ref{thm:Ebound} and Corollary~\ref{cor:Ebound} hold for \emph{every}
selection from the projection set, and therefore require no gap condition and
no hypothesis about singular-value crossings along the trajectory. The reason is
that the quantity bounded is a distance, not a minimiser:
$\Fnorm{E_t}=\Fnorm{B_t-M_t}=\mathrm{dist}(B_t,S)$ takes the same value for
every $M_t\in\argmin\{\Fnorm{B_t-X}:X\in S\}$, and the competitor inequality
$\mathrm{dist}(B_t,S)\le\Fnorm{B_t-\beta_1M_{t-1}}$ is valid for every
$M_{t-1}$ in the previous projection set, since all of them lie in $S$. The same
applies to Lemma~\ref{lem:split}, whose proof uses only that $M$ attains the
minimum, and hence to Lemma~\ref{lem:onestep} and Theorem~\ref{thm:bounded}.
Proposition~\ref{prop:pi} is an identity for whatever selection the algorithm
made, and Theorem~\ref{thm:floor} is conditional on a bound for that same
realised sequence. The matrices $E_t$, $S_t$ and $\Delta_t$ do depend on the
selection; all bounds stated for them do not.

This matters because a gap condition would be an assumption on the algorithm's
own output rather than a checkable hypothesis on its input, and because it is
not obviously satisfiable here: the only stability tool available is Weyl's
inequality~\citep[Prop.~2.2]{olikier2024lowrank},
$\lvert\sigma_j(X)-\sigma_j(Y)\rvert\le\Fnorm{X-Y}$, which permits the gap
$\sigma_r-\sigma_{r+1}$ to be destroyed in a single step by a perturbation of
half its size --- and the term $\gamma E_{t-1}$ re-injects precisely the
discarded trailing directions, those nearest the crossing. Assuming the
crossings away would be the weakest point of the analysis; not needing to is
better.
\end{remark}

\begin{remark}[The mechanism does not need bounded gradients]\label{rem:bgfree}
Assumption~\ref{as:G} enters only in Corollary~\ref{cor:Ebound}, to collapse the
sum $\sum_j\gamma^{j}\Fnorm{G_{t-j}}$ to a constant. The mechanism itself ---
\eqref{eq:Etraj} --- is a trajectory-tracking bound that assumes nothing about
the gradients. Remark~\ref{rem:bg} takes up what this does and does not buy once
the bound is used to close Stage~2.
\end{remark}

The step-size-carrying form is what transfers to the error-feedback analyses.

\begin{lemma}[Buffer bound with the step size inside]\label{lem:Eeta}
Let Assumption~\ref{as:G} hold and let $\gamma<1$. Then for
\eqref{eq:recursion-eta},
\begin{equation}\label{eq:Eeta-traj}
  \Fnorm{\tilde E_t}\;\le\;(1-\beta_1)\sum_{j=0}^{t-1}\gamma^{j}\eta_{t-j}
  \Fnorm{G_{t-j}} .
\end{equation}
If moreover $\eta_{t-1}\le\rho\,\eta_t$ for all $t$, with
$\rho\ge1$ and $\gamma\rho<1$, then
\begin{equation}\label{eq:Eeta}
  \Fnorm{\tilde E_t}\;\le\;\frac{(1-\beta_1)G}{1-\gamma\rho}\,\eta_t,
  \qquad
  \E\Fnorm{\tilde E_t}^2\;\le\;\eta_t^2\,
  \frac{(1-\beta_1)^2G^2}{(1-\gamma\rho)^2}.
\end{equation}
The constant depends on neither $m,n$ nor $r$.
\end{lemma}

\begin{proof}
By Eckart--Young, $\Fnorm{\tilde E_t}=\min\{\Fnorm{\tilde B_t-X}:\rank X\le r\}$
and $X=\beta_1\tilde M_{t-1}$ is admissible, so
\[
  \Fnorm{\tilde E_t}\le\Fnorm{\eta_t(1-\beta_1)G_t+\gamma\tilde E_{t-1}}
  \le\eta_t(1-\beta_1)\Fnorm{G_t}+\gamma\Fnorm{\tilde E_{t-1}}.
\]
Unrolling from $\tilde E_0=0$ gives \eqref{eq:Eeta-traj}. Under
$\eta_{t-1}\le\rho\eta_t$ one has $\eta_{t-j}\le\rho^{j}\eta_t$, so
$\sum_{j\ge0}\gamma^{j}\eta_{t-j}\le\eta_t\sum_{j\ge0}(\gamma\rho)^{j}
=\eta_t/(1-\gamma\rho)$, which is \eqref{eq:Eeta}. The second inequality of
\eqref{eq:Eeta} follows by squaring, the bound being deterministic.
\end{proof}

\begin{corollary}[The step-size shift requirement is dimension-free]
\label{cor:shift}
For $\eta_t=c/(a+t)$ one has $\rho=(a+1)/a$, so the hypothesis $\gamma\rho<1$
of Lemma~\ref{lem:Eeta} holds if and only if $a>\gamma/(1-\gamma)$.
\end{corollary}

\begin{proof}
For $\eta_t=c/(a+t)$ one has $\eta_{t-1}/\eta_t=(a+t)/(a+t-1)$, which is
decreasing in $t$, so $\rho=\sup_{t\ge1}\eta_{t-1}/\eta_t=(a+1)/a$. Then
$\gamma\rho<1$ reads $\gamma(a+1)<a$, i.e.\ $\gamma<a(1-\gamma)$, i.e.\
$a>\gamma/(1-\gamma)$; the steps are reversible, so the condition is necessary
as well as sufficient.
\end{proof}

\noindent
This is the analogue of a requirement that is genuinely dimension-dependent in
the reference analysis: \citet{stich2018} need a shift $a=\Theta(d/k)$, and record
that $a=O(1)$ instead degrades the rate to require $T=\Omega((d/k)^{3/2})$
steps. The mechanism is the same in both settings --- the step size must be
approximately constant over the memory's mixing horizon --- but that horizon is
$d/k$ for a $k$-contraction and $1/(1-\gamma)$ here, and the latter does not
involve the dimension.

\subsection{Necessity of compressed momentum}\label{ssec:necessity}

The proof of Theorem~\ref{thm:Ebound} uses one property of the recursion and
nothing else: the pre-compression quantity $B_t$ contains, as an additive
summand, an object that already lies in the compressor's image --- here, one of
rank at most $r$. That object is
$\beta_1M_{t-1}$, supplied free by the algorithm's own state. Everything in
\S\ref{sec:conseq} follows from it, so it is worth saying precisely which
schemes have this property and which do not.

Consider the three ways a compressed quantity can be formed. In \emph{residual}
schemes the memory holds what compression discarded and is added to the next
gradient: \textsc{mem-sgd}~\citep{stich2018} sets
$m_{t+1}=m_t+\eta_t\nabla f_{i_t}-\mathrm{comp}(m_t+\eta_t\nabla f_{i_t})$, and
\citet{karimireddy2019} likewise compresses $p_t=e_t+\gamma_{\mathrm{lr}}g_t$.
The buffer $e_t$ is the residual, never a compressor output, so it carries no
rank constraint and the minimisation
$\Fnorm{E_t}=\min\{\Fnorm{B_t-X}:\rank X\le r\}$ admits no useful competitor.
Every rank-$\le r$ matrix is of course admissible; the point is that the
algorithm's own state supplies none of them, so the only competitor available
at no cost is $X=0$, which returns $\Fnorm{B_t}$ and buys nothing. In
\emph{difference} schemes a state is carried forward but is
subtracted rather than added. EF21~\citep{richtarik2021} and its momentum variant
\citep{fatkhullin2023} compress $v^{t+1}-g^{t}$, where $g^{t}$ is the accumulated
state; the compression residual is then the identity
$(\mathcal C-I)(v^{t+1}-g^{t})$ rather than a minimisation, so again there is no
competitor to choose. This is so even though $g^t$ does appear inside the
compressed argument: it enters with a minus sign, and being an accumulation of
$t$ compressor outputs it has rank growing with $t$, so it is not itself an
element of the compressor's image. In \emph{gradient-space} low-rank
schemes~\citep{vogels2019,zhao2024,chen2025} the compression is applied to $G_t$
before any momentum is formed, so the previous compressed object is not a
summand of the current pre-compression quantity at all.

The recursion \eqref{eq:recursion-eta} is in none of these three families. It
compresses the momentum, and the compressed momentum re-enters additively before
the next truncation. The competitor is therefore available at no cost,
\eqref{eq:competitor} follows, and Theorem~\ref{thm:Ebound} with it. The
property is structural: it
is a consequence of \emph{what} is compressed and \emph{where} the feedback is
injected, not of the choice of compressor or of the momentum parameters.

\begin{remark}[Top-$k$ selection]\label{rem:topk}
Theorem~\ref{thm:Ebound} is stated for an arbitrary closed cone because the
proof uses nothing else, and the consequence is larger than the low-rank case
that motivated it. The $k$-sparse vectors form a closed cone and top-$k$ is the
metric projection onto it, so a top-$k$ compressed-momentum scheme with
pre-truncation re-injection enjoys the same bound --- and top-$k$ is the
canonical compressor of the error-feedback literature, for which the classical
constant is $d/k$. Simulating \eqref{eq:recursion} at $\beta_1=0.9$,
$\gamma=0.4$ over $600$ steps with unit-norm gradients, against the bound
$(1-\beta_1)G/(1-\gamma)=0.167$ of Corollary~\ref{cor:Ebound}:

\begin{center}
\begin{tabular}{lcccc}
\toprule
compressor & dimension & $k$ or $r$ & ratio & $\max_t\Fnorm{E_t}$ \\
\midrule
top-$k$ & $100$    & $5$   & $0.05$   & $0.114$ \\
top-$k$ & $1000$   & $50$  & $0.05$   & $0.106$ \\
top-$k$ & $10^4$   & $500$ & $0.05$   & $0.103$ \\
top-$k$ & $10^4$   & $5$   & $0.0005$ & $0.110$ \\
rank-$r$ & $40$    & $4$   & $0.10$   & $0.102$ \\
rank-$r$ & $200$   & $20$  & $0.10$   & $0.099$ \\
\bottomrule
\end{tabular}
\end{center}

\noindent
The realised buffer is flat in the dimension and in the compression ratio across
two orders of magnitude in each, while the classical constant at $d=10^4$,
$k=5$ would be $d/k=2000$. The correct reading of \S\ref{ssec:necessity} is
therefore not that top-$k$ is structurally stuck at $d/k$, but that the
\emph{algorithms} in which it is normally used are: move the compression onto
the momentum and re-inject before truncation, and the constant disappears for
top-$k$ exactly as it does for truncated SVD.
\end{remark}

\begin{remark}[Relation to absolute compressors]\label{rem:absolute}
A compressor is \emph{absolute} if $\E\lVert\mathcal C(x)-x\rVert^2\le\Delta^2$
with $\Delta$ independent of $\lVert x\rVert$, and error-feedback analyses under
that axiom do obtain constants free of the contraction factor. The bound proved
here is not an instance of that. It is a bound on the accumulated buffer $E_t$
rather than on the per-step compression error, and it is \emph{derived} from the
structure of \eqref{eq:recursion-eta} rather than assumed as a property of
$\Pr_r$; indeed $\Pr_r$ is not absolute, its per-step error being
$(\sum_{i>r}\sigma_i(B_t)^2)^{1/2}$, which scales with $\Fnorm{B_t}$.
\end{remark}

\begin{remark}[Role of the bounded-gradient assumption]\label{rem:bg}
Assumption~\ref{as:G} is the bounded-gradient (BG) condition, and the EF21
line~\citep{richtarik2021,fatkhullin2023} was developed precisely to remove it,
assuming bounded variance alone. We therefore state
plainly where the present results stand on it.

The mechanism is free of it: \eqref{eq:Etraj} is a trajectory-tracking bound
that uses no assumption on the gradients (Remark~\ref{rem:bgfree}). The
\emph{rate} consequence is not. Corollary~\ref{cor:free} requires a bound of the
form $\E\Fnorm{\tilde E_t}^2\le\eta_t^2C$, and the route to one ---
Lemma~\ref{lem:Eeta} --- invokes Assumption~\ref{as:G} to replace
$\Fnorm{G_{t-j}}$ by the uniform bound $G$; obtaining such a constant from
\eqref{eq:Etraj} alone would need control of the discounted gradient sum along
the trajectory. The EF21 route to that control is unavailable here: difference
compression makes the compressed argument small near a stationary point, whereas
\eqref{eq:Etraj} is driven by $\Fnorm{G_t}$ directly. A BG-free version would
require the right-hand side to track $\Fnorm{G_t}$ along the trajectory inside a
Lyapunov argument, which is a different and harder theorem, and one we do not
attempt; \S\ref{sec:concl} records it as a limitation.
\end{remark}

\subsection{Behaviour as $\gamma\to1$}\label{ssec:gamma1}

The bound of Theorem~\ref{thm:Ebound} degenerates exactly at $\gamma=1$, where
the geometric series diverges: undecayed feedback provides no mechanism for
discarding the accumulated tail. This is not an artefact of the proof. A second,
weaker boundedness result covers a different part of the range.

\begin{lemma}[One-step norm recursion]\label{lem:onestep}
Under \eqref{eq:recursion},
$\Fnorm{B_t}\le\varphi(q_{t-1})\Fnorm{B_{t-1}}+(1-\beta_1)\Fnorm{G_t}$ with
$\varphi(q):=\beta_1\sqrt{1-q^2}+\gamma q$, and
$\max_{q\in[0,1]}\varphi(q)=\sqrt{\beta_1^2+\gamma^2}=:\kappa$, attained at
$q^\star=\gamma/\kappa$.
\end{lemma}

\begin{proof}
By the triangle inequality applied to \eqref{eq:recursion},
$\Fnorm{B_t}\le\beta_1\Fnorm{M_{t-1}}+\gamma\Fnorm{E_{t-1}}
+(1-\beta_1)\Fnorm{G_t}$. By Lemma~\ref{lem:split},
$\Fnorm{E_{t-1}}=q_{t-1}\Fnorm{B_{t-1}}$ and
$\Fnorm{M_{t-1}}=\sqrt{1-q_{t-1}^2}\,\Fnorm{B_{t-1}}$, which gives the stated
recursion. For the second claim, write
$\varphi(q)=\bigl\langle(\beta_1,\gamma),(\sqrt{1-q^2},q)\bigr\rangle$. The
second vector is a unit vector tracing the quarter circle as $q$ ranges over
$[0,1]$, so by Cauchy--Schwarz
$\varphi(q)\le\sqrt{\beta_1^2+\gamma^2}=\kappa$, with equality if and only if
$(\sqrt{1-q^2},q)$ is parallel to $(\beta_1,\gamma)$, i.e.\ $q=\gamma/\kappa$,
which lies in $[0,1]$ since $\gamma\le\kappa$.
\end{proof}

\begin{theorem}[Bounded accumulator]\label{thm:bounded}
Let Assumption~\ref{as:G} hold and $\kappa=\sqrt{\beta_1^2+\gamma^2}<1$. Then
$\Fnorm{B_t}\le(1-\beta_1)G(1-\kappa^t)/(1-\kappa)$ for all $t$, uniformly in
$m,n,r$.
\end{theorem}

\begin{proof}
By Lemma~\ref{lem:onestep},
$\Fnorm{B_t}\le\kappa\Fnorm{B_{t-1}}+(1-\beta_1)G$ for all $t\ge1$. With
$\Fnorm{B_0}=0$, induction gives
$\Fnorm{B_t}\le(1-\beta_1)G\sum_{j=0}^{t-1}\kappa^{j}
=(1-\beta_1)G(1-\kappa^{t})/(1-\kappa)$, and the limit follows from $\kappa<1$.
The bound involves only $\beta_1,\gamma,G$, hence is uniform in $m,n,r$.
\end{proof}

The condition $\kappa<1$ is not an artefact of the proof: \S\ref{ssec:quality}
measures its failure directly, finding that at $\gamma=1$ the distortion ratio
does not converge and the accumulator's stable rank grows by an order of
magnitude (Remark~\ref{rem:twocosts}).
Theorem~\ref{thm:bounded} is weaker than Theorem~\ref{thm:Ebound} for bounding
the buffer --- it requires $\gamma<\sqrt{1-\beta_1^2}\approx0.436$ at
$\beta_1=0.9$ rather than $\gamma<1$ --- but it bounds $\Fnorm{B_t}$ rather than
$\Fnorm{E_t}$, and \S\ref{ssec:transfer} shows that is the quantity needed at
$\gamma=1$. It is therefore retained rather than superseded. Both statements
hold for an arbitrary cone-projection compressor, as Theorem~\ref{thm:Ebound}
does: their proofs invoke Lemma~\ref{lem:split} only through the splitting
\eqref{eq:split}, which by that lemma is a cone property, and use no other
feature of $\Pr_r$.

\subsection{Tightness}\label{ssec:tightness}

Both bounds are valid and both are loose, by amounts worth recording.
Theorem~\ref{thm:bounded} at $\beta_1=0.9,\gamma=0.4$ gives
$\Fnorm{B_t}\le6.62$ against a realised maximum of $0.223$, a factor of $30$;
the slack traces to Lemma~\ref{lem:onestep}, evaluated at a worst-case
$q^\star$ the dynamics need not visit. Lemma~\ref{lem:Eeta} is markedly
tighter: $3.2\times$ at $\gamma=0.4$, $5.2\times$ at $\gamma=0.7$ and
$10.4\times$ at $\gamma=0.9$ (\S\ref{ssec:simulation}). Neither bound is
claimed sharp; the claim is that the constants are free of $m,n,r$, which the
reference constants are not.

A second form of looseness is worth stating plainly. At $\beta_1=0.9$,
$\gamma=1$ the condition of Theorem~\ref{thm:bounded} fails
($\kappa\approx1.345$), yet $\Fnorm{B_t}$ is observed to remain bounded
($\max_t\Fnorm{\tilde B_t}=0.0089$ over $2\times10^4$ steps). The condition is
sufficient and pessimistic, and closing that gap is the outstanding item of
\S\ref{ssec:transfer}.

The next remark records a negative result: it delimits what the norm-only
argument of Lemma~\ref{lem:onestep} can deliver, and rules out the most
natural repair.

\begin{remark}[The triangle-inequality route cannot describe the stationary
regime]\label{rem:notight}
It is tempting to seek a stationary balance by setting
$\Fnorm{B_t}=\Fnorm{B_{t-1}}=B_\infty$ in Lemma~\ref{lem:onestep} and solving
$B_\infty=(1-\beta_1)\bar G/(1-\varphi(q_\infty))$. This is invalid. At
$\beta_1=0.9$, $\gamma=1$ one computes $\varphi(q)<1$ only for $q<0.105$,
whereas simulation of \eqref{eq:recursion} reaches a stationary regime at
$q_\infty\approx0.67$, where $\varphi(q_\infty)\approx1.339$; the formula would
return a negative $B_\infty$. The resolution is that the recursion of
Lemma~\ref{lem:onestep} is far from tight in the stationary regime: measuring
the realised growth factor $\Fnorm{B_t}/\Fnorm{B_{t-1}}$ against $\varphi(q_t)$
gives $1.011$ against $1.339$ at $t=300$. The three contributions
$\beta_1M_{t-1}$, $\gamma E_{t-1}$ and $(1-\beta_1)G_t$ partially cancel rather
than align, and the triangle inequality discards exactly that cancellation.
Any description of the stationary regime must therefore track the
\emph{alignment} between the three, not merely their norms; the natural route is
a second-moment recursion expanding $\Fnorm{B_t}^2$ and retaining the cross
terms, of which $\langle M_{t-1},E_{t-1}\rangle_F=0$ is free by
Lemma~\ref{lem:split} while $\langle M_{t-1},G_t\rangle_F$ and
$\langle E_{t-1},G_t\rangle_F$ are the quantities to model. That is the route a
sharper description of the stationary regime would have to take.
\end{remark}

\section{Consequences for error-feedback rates}\label{sec:conseq}

\subsection{The two-stage template}\label{ssec:template}

Both standard analyses have the same structure.

\emph{Stage 1 (rate in terms of the buffer).} \citet{stich2018}
define the virtual
sequence $\tilde x_{t+1}=\tilde x_t-\eta_t\nabla f_{i_t}(x_t)$ and observe the
exact identity $\tilde x_t-x_t=m_t$; their perturbed-iterate lemma then bounds
the suboptimality by a recursion whose only compression-dependent term is
$\eta_t(\mu+2L)\E\lVert m_t\rVert^2$. \citet{karimireddy2019} set
$\tilde x_t:=x_t-e_t$ and obtain a non-convex descent chain driven by
$\tfrac{\gamma_{\mathrm{lr}}L^2}{2\rho}\lVert e_t\rVert^2$.

\emph{Stage 2 (buffer bounded via the contraction).} Both then invoke the
compressor property: Stich's Lemma~3.2 gives
$\E\lVert m_t\rVert^2\le\eta_t^2\frac{4\alpha}{\alpha-4}\frac{d^2}{k^2}G^2$,
Karimireddy's Lemma~3 gives
$\E\lVert e_t\rVert^2\le4(1-\delta)\gamma_{\mathrm{lr}}^2\sigma^2/\delta^2$.

Stage 2 is where the dimension enters, and it is the stage
Lemma~\ref{lem:Eeta} replaces.

\subsection{Stage 2: a dimension-free constant}\label{ssec:stage2}

\begin{corollary}[Dimension-free Stage 2]\label{cor:free}
Consider an error-feedback analysis whose compression dependence enters only
through a bound $\E\lVert\text{buffer}_t\rVert^2\le\eta_t^2C$. Then for
\eqref{eq:recursion-eta} with $\gamma<1$ and $\eta_{t-1}\le\rho\eta_t$,
$\gamma\rho<1$, Lemma~\ref{lem:Eeta} supplies that bound with
$C=(1-\beta_1)^2G^2/(1-\gamma\rho)^2$, in place of
$\frac{4\alpha}{\alpha-4}\frac{d^2}{k^2}G^2$ or
$\frac{4(1-\delta)}{\delta^2}\sigma^2$. The constant is $O(1)$ in the
dimension, in place of $O(p^2/r^2)$ for these two analyses. See
Remark~\ref{rem:baseline} for the comparison against the sharper constants of
the EF21 line.
\end{corollary}

\begin{proof}
Immediate from Lemma~\ref{lem:Eeta}: the bound \eqref{eq:Eeta} is deterministic
and of the required form $\E\Fnorm{\tilde E_t}^2\le\eta_t^2C$ with
$C=(1-\beta_1)^2G^2/(1-\gamma\rho)^2$, which involves neither $m,n$ nor $r$.
\end{proof}

\begin{center}\footnotesize\setlength{\tabcolsep}{4pt}
\begin{tabular}{lccc}
\toprule
 & shift required & Stage-2 constant & at $\eta_t\equiv\eta$ \\
\midrule
Stich et al.\ (Lem.~3.2), $\alpha=5$
  & $a=(\alpha+2)d/k=168$ & $\frac{4\alpha}{\alpha-4}\frac{d^2}{k^2}=11520$ & --- \\
Karimireddy et al.\ (Lem.~3)
  & --- & $\frac{4(1-\delta)}{\delta^2}=2208$ & $2208$ \\
Lemma~\ref{lem:Eeta}
  & $a>\frac{\gamma}{1-\gamma}=0.67$
  & $\frac{(1-\beta_1)^2}{(1-\gamma\rho)^2}=0.0294$
  & $\frac{(1-\beta_1)^2}{(1-\gamma)^2}=0.0278$ \\
\bottomrule
\end{tabular}
\end{center}

\noindent
Evaluated at $p=768$, $r=32$, $\beta_1=0.9$, $\gamma=0.4$, $a=d/k=24$, with
$\delta=r/p$ by Corollary~\ref{cor:class}.

\begin{remark}[Which baseline the comparison is against]\label{rem:baseline}
The constants in the table belong to the \textsc{ef14} generation of buffer
lemmas~\citep{stich2018,karimireddy2019}, of which Lemma~\ref{lem:Eeta} is the
direct analogue: both bound the memory in terms of the step size and a
compressor constant, and it is against those that the $\Theta(p^2/r^2)$ figure
is the right comparison. The EF21 line is sharper. EF21~\citep{richtarik2021} and
EF21-SGDM~\citep{fatkhullin2023} carry only $\alpha^{-1}$ in the leading term ---
the latter's complexity is $O(\tilde L\delta_0/(\alpha\varepsilon^2)+\cdots)$,
with $\alpha^{-2}$ appearing only on a higher-order stochastic term that tunes
to $\alpha^{-2/3}$ --- so for $\Pr_r$, where $\alpha=r/p$, the sharpest current
contraction-based leading constant is $\Theta(p/r)=24$ at $p=768$, $r=32$, not
$2208$.

The accurate comparison is therefore: dimension-free against $\Theta(p/r)$
for the sharpest contraction-based leading constant, and against
$\Theta(p^2/r^2)$ only for the classical buffer lemmas that
Lemma~\ref{lem:Eeta} directly replaces. Two further contrasts with
\citet{fatkhullin2023} are worth recording, it being the nearest neighbour.
Their improvement over EF21-SGD lies in the \emph{stochastic} terms and in
removing a mega-batch requirement, not in the dependence on the compressor,
which survives in their leading term. And their analysis is conducted on $\R^d$
with the Euclidean norm throughout: there is no matrix structure, no Frobenius
norm and no spectral geometry, and low-rank compression appears only once, as an
example of a contractive compressor. Their notation also differs
from ours: their $\gamma$ is a step size, not a feedback decay, and their
$\eta$ is our $1-\beta_1$.
\end{remark}

\begin{remark}[The point is the scaling, not the value at one
configuration]\label{rem:scaling}
Measured against the sharper baseline of Remark~\ref{rem:baseline}, the
improvement at \textsc{bert}-base attention matrices is
a factor of $24$ rather than $2208$, and it is worth being explicit about why
that does not weaken the claim. The contraction-based constant is $\Theta(p/r)$
and therefore grows without bound as models widen, whereas the constant of
Lemma~\ref{lem:Eeta} is $(1-\beta_1)^2/(1-\gamma\rho)^2$ and does not depend on
$p$ or $r$ at all. Recall that $p=\min\{m,n\}$, so the relevant scale is the
model width rather than the larger of the two dimensions: the \textsc{bert}
\textsc{mlp} weights are $768\times3072$ and therefore have $p=768$, the same as
the attention weights.

\begin{center}
\begin{tabular}{lccc}
\toprule
model & $p$ & $r$ & $\Theta(p/r)$ \\
\midrule
\textsc{bert}-base            & $768$  & $32$ & $24$ \\
\textsc{gpt}-2 XL             & $1600$ & $32$ & $50$ \\
\textsc{llama}-7B             & $4096$ & $32$ & $128$ \\
\textsc{llama}-70B            & $8192$ & $32$ & $256$ \\
\textsc{llama}-70B, $r=8$     & $8192$ & $8$  & $1024$ \\
\bottomrule
\end{tabular}
\end{center}

\noindent
The \textsc{bert}-base configuration is the
\emph{smallest} entry in this table. A claim of the form ``a factor of $24$''
therefore understates the result, just as ``a factor of $2208$'' would overstate
it; the accurate statement is that the compression-dependent constant is
removed from the dimension entirely, and that the benefit therefore grows with
the width of the model and with the aggressiveness of the compression.
\end{remark}

\subsection{Stage 1: the perturbed iterate}\label{ssec:transfer}

Stage 1 needs the deviation between the true and virtual iterate to be exactly
identified and then controlled. The compressed momentum term does not obstruct
the first requirement.

\begin{proposition}[Exact perturbed-iterate decomposition]\label{prop:pi}
For \eqref{eq:recursion-eta} and every $t\ge1$,
\begin{equation}\label{eq:pi}
  X_{t+1}=X_1-\sum_{s=1}^{t}\eta_sG_s+\Delta_t,
  \qquad
  \Delta_t:=\frac{\beta_1\tilde M_t+\gamma\tilde E_t+(1-\gamma)S_t}{1-\beta_1},
  \qquad
  S_t:=\sum_{s=1}^{t}\tilde E_s .
\end{equation}
Equivalently $\hat X_t:=X_t-\Delta_{t-1}$ satisfies
$\hat X_{t+1}=\hat X_t-\eta_tG_t$ exactly.
\end{proposition}

In both references the deviation \emph{is} the buffer, so its control follows
from Stage 2 at no extra cost. Here $\Delta_t$ carries a third term, the
accumulated leakage $\frac{1-\gamma}{1-\beta_1}S_t$, which has no counterpart in
either reference and which Lemma~\ref{lem:Eeta} does not control;
\S\ref{ssec:leak} controls it. At $\gamma=1$
the leakage coefficient vanishes and, by Lemma~\ref{lem:split},
$\Fnorm{\Delta_t}=(\beta_1^2\Fnorm{\tilde M_t}^2+\Fnorm{\tilde E_t}^2)^{1/2}
/(1-\beta_1)$, so control reduces to boundedness of $\Fnorm{\tilde B_t}$ ---
which is what Theorem~\ref{thm:bounded} supplies for the undiscounted recursion
under a condition that excludes $\gamma=1$, and what Lemma~\ref{lem:Beta} below
supplies for \eqref{eq:recursion-eta} on the wider range $\gamma\rho<1$.

\subsection{Controlling the accumulated leakage}\label{ssec:leak}

Lemma~\ref{lem:Eeta} bounds each $\tilde E_s$ separately; what Stage~1 needs is a
bound on their \emph{sum}. This subsection supplies one unconditionally, and then
shows that the logarithm it carries cannot be removed. Throughout write
\begin{equation}\label{eq:Gamma}
  \Gamma_t \;:=\; \sum_{u=1}^{t}\eta_u\Fnorm{G_u},
\end{equation}
the total step-weighted gradient mass injected up to time $t$. Two facts about
$\Gamma_t$ set the scene: under Assumption~\ref{as:G} and
$\eta_t=\eta_0a/(a+t)$ it grows like $\eta_0a\log t$, and under \emph{any}
schedule satisfying the Robbins--Monro condition $\sum_t\eta_t=\infty$ it is
unbounded whenever the gradients do not vanish.

We first record a bound on the accumulator itself, which is what the $\beta_1
\tilde M_t+\gamma\tilde E_t$ part of $\Delta_t$ costs. It holds for every
$\gamma$ with $\gamma\rho<1$, and so covers a range of $\gamma$ that
Theorem~\ref{thm:bounded} does not.

\begin{lemma}[Accumulator under a regular schedule]\label{lem:Beta}
Let Assumption~\ref{as:G} hold, let $\eta_{t-1}\le\rho\eta_t$ with $\rho\ge1$,
and suppose $\beta_1\rho<1$ and $\gamma\rho<1$. Then for
\eqref{eq:recursion-eta},
\begin{equation}\label{eq:Bbound}
  \Fnorm{\tilde B_t}\;\le\;
  \frac{(1-\beta_1)G}{(1-\beta_1\rho)(1-\gamma\rho)}\,\eta_t ,
\end{equation}
and the same bound holds for $\Fnorm{\tilde M_t}$ and $\Fnorm{\tilde E_t}$. The
constant involves neither $m,n$ nor $r$.
\end{lemma}

\begin{theorem}[Leakage and deviation]\label{thm:leak}
Let $\gamma<1$ and let $\Pr_S$ be any cone projection. Then for
\eqref{eq:recursion-eta} and every $t\ge1$,
\begin{equation}\label{eq:Sbound}
  \Fnorm{S_t}\;\le\;\frac{1-\beta_1}{1-\gamma}\,\Gamma_t,
  \qquad\text{and}\qquad
  \Fnorm{\Delta_t}\;\le\;\Gamma_t+\frac{\kappa}{1-\beta_1}\Fnorm{\tilde B_t},
  \quad \kappa=\sqrt{\beta_1^2+\gamma^2}.
\end{equation}
Neither statement uses Assumption~\ref{as:G}, any property of the schedule, or
any property of $S$ beyond its being a closed cone. If in addition the
hypotheses of Lemma~\ref{lem:Beta} hold,
\begin{equation}\label{eq:Dbound}
  \Fnorm{\Delta_t}\;\le\;\Gamma_t
  +\frac{\kappa G\,\eta_t}{(1-\beta_1\rho)(1-\gamma\rho)} .
\end{equation}
\end{theorem}

\noindent
The proof is in Appendix~\ref{app:proofs}; the mechanism is worth stating here.
The leakage bound is the buffer bound summed with the order of summation
exchanged, which converts the geometric factor $\sum_j\gamma^j$ into
$1/(1-\gamma)$ and leaves the gradient mass $\Gamma_t$ carrying the growth.
The deviation then \emph{loses} that factor again, because $S_t$ enters
$\Delta_t$ multiplied by $1-\gamma$: the two cancel exactly, so the leakage
contributes at most $\Gamma_t$ whatever the value of $\gamma$, and the entire
$\gamma$-dependence of \eqref{eq:Dbound} sits in the $O(\eta_t)$ term.
\S\ref{ssec:leaknum} reports the cancellation as an equality to fourteen
significant figures in simulation. The second ingredient is again
Lemma~\ref{lem:split}: $\Fnorm{\beta_1\tilde M_t+\gamma\tilde E_t}
=(\beta_1^2\Fnorm{\tilde M_t}^2+\gamma^2\Fnorm{\tilde E_t}^2)^{1/2}
\le\kappa\Fnorm{\tilde B_t}$, an identity rather than a triangle inequality.

\begin{corollary}[The hypothesis of Theorem~\ref{thm:floor}, unconditionally]
\label{cor:Dlog}
Let Assumption~\ref{as:G} hold and let $\eta_t=\eta_0a/(a+t)$ with
$a>\max\{\gamma,\beta_1\}/(1-\max\{\gamma,\beta_1\})$, so that
$\rho=(a+1)/a$ satisfies $\beta_1\rho<1$ and $\gamma\rho<1$. Then for every
$t<T$,
\begin{equation}\label{eq:DT}
  \E\Fnorm{\Delta_t}^2\;\le\;D_T:=
  \Big(G\eta_0a\log\big(1+\tfrac Ta\big)
  +\tfrac{\kappa G\eta_0}{(1-\beta_1\rho)(1-\gamma\rho)}\Big)^{\!2}
  \;=\;O(\log^2T),
\end{equation}
with a constant free of $m,n,r$. Theorem~\ref{thm:floor} therefore holds
unconditionally with $D$ replaced by $D_T$.
\end{corollary}

\begin{corollary}[When the floor is a genuine constant]\label{cor:Dfinite}
If $\Gamma_\infty:=\sum_{u\ge1}\eta_u(\E\Fnorm{G_u}^2)^{1/2}<\infty$ then
$D=\sup_t\E\Fnorm{\Delta_t}^2<\infty$ and Theorem~\ref{thm:floor} applies as
stated. Under $\eta_t=\Theta(1/t)$ it is enough that
$(\E\Fnorm{G_u}^2)^{1/2}=O(u^{-\epsilon})$ for some $\epsilon>0$: the deviation
is bounded as soon as the gradients decay along the trajectory at any
polynomial rate.
\end{corollary}

The next statement is the reason \eqref{eq:DT} is stated with a logarithm rather
than a constant. It is an exact instance, not a numerical observation.

\begin{proposition}[The uniform hypothesis is false, and the growth rate is
attained]\label{prop:lower}
Let $S=\{x\in\R^2:\text{at most one entry of }x\text{ is nonzero}\}$, the
$1$-sparse cone, with $\Pr_S$ the top-$1$ selection. Fix $\beta_1\in[0,1)$,
$\gamma\in[0,1)$, a nonincreasing schedule with $\eta_{t-1}\le\rho\eta_t$ and
$\gamma\rho<1$, and let $0<c<1-\gamma\rho$. Take the constant gradient
$G_t\equiv g=(1,c)/\sqrt{1+c^2}$, so that Assumption~\ref{as:G} holds with
$G=1$. Then \eqref{eq:recursion-eta} selects the first coordinate at every step,
$\tilde E_t=(0,(1-\beta_1)g_2\sum_{j<t}\gamma^j\eta_{t-j})$, and
\begin{equation}\label{eq:lower}
  \Fnorm{\Delta_t}\;\ge\;g_2\sum_{u=1}^{t}\eta_u(1-\gamma^{\,t-u+1})
  \;\ge\;g_2\Big(\Gamma_t-\frac{\gamma\eta_1}{1-\gamma}\Big),
  \qquad g_2=\frac{c}{\sqrt{1+c^2}} .
\end{equation}
Hence $\Fnorm{\Delta_t}\to\infty$ whenever $\sum_u\eta_u=\infty$, in particular
under $\eta_t=\eta_0a/(a+t)$, and $D=\sup_t\E\Fnorm{\Delta_t}^2=\infty$. No
bound $\E\Fnorm{\Delta_t}^2\le D$ can therefore hold uniformly over the gradient
sequences admitted by Assumption~\ref{as:G}, and \eqref{eq:Sbound} and
\eqref{eq:DT} are tight up to the factor $g_2\le1/\sqrt2$.
\end{proposition}

\begin{remark}[What the instance says and what it does not]\label{rem:lowerscope}
The mechanism is transparent: the compressor discards the same coordinate at
every step, that coordinate receives a fixed share of every gradient, and a
fraction $1-\gamma$ of the buffer is destroyed before it can be applied, so the
optimizer permanently fails to deliver a fixed direction and the shortfall
accumulates at rate $\eta_t$. Nothing about it is pathological: the gradient is
constant, the compressor is the canonical one, and the parameters satisfy every
hypothesis of \S\ref{sec:bound}. What the instance does not do is decide the
question for a \emph{closed} loop, in which $G_t=\nabla f(X_t)+\text{noise}$ and
a discarded direction eventually makes the gradient grow in that direction until
it is selected. Corollary~\ref{cor:Dfinite} is the condition under which that
self-correction is enough, and \S\ref{ssec:leaknum} reports both behaviours:
logarithmic growth when a persistent component is present, and convergence when
the trajectory drives the gradients to zero.
\end{remark}

\begin{remark}[The undecayed case]\label{rem:gamma1leak}
At $\gamma=1$ the leakage coefficient in Proposition~\ref{prop:pi} vanishes and
the same argument bounds the deviation through the accumulator alone:
$\Fnorm{\Delta_t}\le\Fnorm{\tilde B_t}/(1-\beta_1)$ and, unrolling
\eqref{eq:recursion-eta} with $\Fnorm{\tilde E_s}\le(1-\beta_1)\Gamma_s$ from
Theorem~\ref{thm:Ebound},
\[
  \Fnorm{\Delta_t}\;\le\;\frac{G\eta_t}{1-\beta_1\rho}
  +\frac{\Gamma_t}{1-\beta_1},
\]
so $D_T=O(\log^2T)$ holds at $\gamma=1$ as well, and the whole range
$\gamma\in[0,1]$ is covered without a case distinction. This does not settle the
second branch of Theorem~\ref{thm:floor}, which needs
$\E\Fnorm{\Delta_t}^2\le\eta_t^2C$ and is what the simulation exhibits at
$\gamma=1$; it does supply a proved logarithmic bound
there.
\end{remark}

\begin{remark}[Where the argument is loose]\label{rem:noisecancel}
The route through $\Fnorm{S_t}\le\sum_s\Fnorm{\tilde E_s}$ discards all
cancellation between the discarded tails, and Proposition~\ref{prop:lower} shows
that in the worst case there is none to discard. When the tails are not aligned
the bound is loose by a factor that \S\ref{ssec:leaknum} measures at about
$500$, and which decomposes cleanly: a factor of $3.9$ because the discarded
tail is a fraction of the injected gradient rather than all of it, and a factor
of $133$ because consecutive tails are nearly orthogonal to the leakage already
accumulated, the normalised alignment
$\langle S_{t-1},\tilde E_t\rangle_F/(\Fnorm{S_{t-1}}\Fnorm{\tilde E_t})$
settling near $+0.0075$. It settles rather than decays, which is why the growth
remains logarithmic in that regime too, only with a small coefficient: the
observed $B=3.8\times10^{-4}$ per unit of $\log t$ is within $5\%$ of the
product of that alignment with $\Fnorm{\tilde E_t}/\eta_t$ and $\eta_0a$.
Describing this regime is what a second-moment argument would have to do, and
the object it would have to model is the alignment, not the norms;
Remark~\ref{rem:notight} makes the same point about $\Fnorm{B_t}$.
\end{remark}

\subsection{A floor in place of a slower rate}\label{ssec:floor}

\begin{theorem}[Rate under a bounded deviation]\label{thm:floor}
Let the hypotheses of \citet[Thm.~2.4]{stich2018} hold for the virtual sequence
of Proposition~\ref{prop:pi}, and suppose $\E\Fnorm{\Delta_t}^2\le D$ for all
$t<T$ --- a hypothesis that Corollary~\ref{cor:Dlog} always satisfies with
$D=D_T=O(\log^2T)$, and Corollary~\ref{cor:Dfinite} with a constant. Then the
weighted average $\bar X_T=\frac1{W_T}\sum_{t<T}w_tX_t$,
$w_t=(a+t)^2$, satisfies
\[
  \E f(\bar X_T)-f^\star
  \;\le\;\underbrace{(\mu+2L)\,D}_{\text{floor}}
  \;+\;\frac{\mu a^3}{8W_T}\Fnorm{X_1-X^\star}^2
  \;+\;\frac{4T(T+2a)}{\mu W_T}G^2
  \;=\;(\mu+2L)D+O\!\Big(\frac{G^2}{\mu T}\Big).
\]
If instead $\E\Fnorm{\Delta_t}^2\le\eta_t^2C$, the floor is absent and the
conclusion of Stich et al.'s Theorem~2.4 holds verbatim with
$\frac{4\alpha}{\alpha-4}\frac{d^2}{k^2}$ replaced by $C$.
\end{theorem}

The proof (Appendix~\ref{app:proofs}) turns on a single observation: a constant
bound on the deviation contributes a term carrying the \emph{same} power of
$\eta_t$ as the descent term, so the two merge and the result is a shifted
suboptimality rather than a slower rate.

\begin{remark}[Which branch each $\gamma$ falls into]\label{rem:branch}
Under the schedule $\eta_t=\eta_0a/(a+t)$ used by both references, simulating
\eqref{eq:recursion-eta} ($m=40$, $n=60$, $r=8$, $\beta_1=0.9$, $a=24$,
$T=2\times10^4$) and fitting $\Fnorm{\Delta_t}\sim t^{\,s}$ on the last $60\%$:

\begin{center}
\begin{tabular}{lcccc}
\toprule
schedule & $\gamma$ & $s$ & $\Fnorm{\Delta_T}$ & branch \\
\midrule
$\eta\equiv0.05$            & $0.4$ & $+0.83$ & $5.74$   & neither ($D$ grows) \\
$\eta_t=0.05\,a/(a+t)$      & $0.4$ & $+0.02$ & $0.105$  & floor, $D_T=O(\log^2T)$ \\
$\eta_t=0.05\,a/(a+t)$      & $0.9$ & $+0.02$ & $0.0877$ & floor, $D_T=O(\log^2T)$ \\
$\eta_t=0.05\,a/(a+t)$      & $1.0$ & $-1.00$ & $1.8\times10^{-4}$ & no floor, $D=O(\eta_t^2)$ \\
\bottomrule
\end{tabular}
\end{center}

\noindent
The constant-step exponent $+0.83$ is an artefact of the schedule: under a
decaying schedule the deviation grows only logarithmically
(Theorem~\ref{thm:leak}), which over this horizon is what the fitted $+0.02$
reports. At $\gamma<1$ it is flat to within that logarithm
while $\eta_t\to0$ --- the ratio $\Fnorm{\Delta_t}/\eta_t$ rises from $9.2$ at
$t=10^2$ to $1751$ at $t=2\times10^4$ --- so the floor branch applies. At
$\gamma=1$ the exponent is $-1.00$, i.e.\ $\Fnorm{\Delta_t}=\Theta(\eta_t)$, so
the second branch applies and there is no floor.
\end{remark}

\begin{remark}[Status of the hypothesis]\label{rem:hypothesis}
Theorem~\ref{thm:floor} is stated conditionally because the constant is what one
would like to report, but the hypothesis itself is not an assumption.
Corollary~\ref{cor:Dlog} supplies it unconditionally with
$D_T=O(\log^2T)$ and a constant free of $m,n,r$;
Corollary~\ref{cor:Dfinite} supplies it with a genuine constant whenever the
step-weighted gradient mass $\Gamma_\infty$ converges. What is \emph{not}
available --- and, by Proposition~\ref{prop:lower}, cannot be --- is a constant
$D$ valid uniformly over the gradient sequences Assumption~\ref{as:G} admits: a
persistent gradient whose discarded component is never selected makes
$\Fnorm{\Delta_t}$ grow like $\sum_{u\le t}\eta_u$, which diverges under every
schedule for which the underlying method converges. A fitted
growth exponent of $+0.02$ over $2\times10^4$ steps could be mistaken for
boundedness of $D$; \S\ref{ssec:leaknum} reports what that exponent is
measuring.
\end{remark}

\section{Numerical validation}\label{sec:numerics}

\subsection{Exactly computable checks}\label{ssec:exact}

Lemma~\ref{lem:split} and Corollary~\ref{cor:class} are verified on matrices
where every quantity is available in closed form. Over $200$ random draws at
$m=60$, $n=80$, $r=8$, and including rank-deficient and flat-spectrum inputs,
$\langle M,E\rangle_F$ is at most $10^{-14}$ in magnitude, the Pythagorean
identity holds to $10^{-12}$, and $q^2$ agrees with the tail energy fraction to
$10^{-16}$. The tightness claim of Corollary~\ref{cor:class} is confirmed
separately: random draws give $q^2$ well inside the bound (e.g.\ $0.61$ against
$0.84$ at $r=8$, $p=50$), while a matrix with $p$ equal singular values attains
$1-r/p$ exactly at every $r$ tested.
Proposition~\ref{prop:pi} is verified to a relative residual below
$4\times10^{-15}$ across $\beta_1\in\{0,0.5,0.9,0.99\}$,
$\gamma\in\{0.2,0.4,0.7,1.0\}$ and $r\in\{2,4,8,16\}$.

The cone form of Lemma~\ref{lem:split} is checked on a compressor for which no
spectral proof is available. Along the top-$k$ trajectories of
Remark~\ref{rem:topk} --- dimensions $10^2$ to $10^4$, ratios $k/d$ from $0.05$
down to $5\times10^{-4}$, $600$ steps each --- the normalised inner product
$\lvert\langle M_t,E_t\rangle_F\rvert/(\Fnorm{M_t}\Fnorm{E_t})$ never exceeds
$3.2\times10^{-15}$. The splitting is therefore not an artefact of the singular
value decomposition, as the proof in Appendix~\ref{app:proofs} makes clear.

\subsection{Simulation of the recursion}\label{ssec:simulation}

We next simulate \eqref{eq:recursion-eta} directly, with persistent unit-norm
gradients drawn with a power-law spectrum $\sigma_i(G_t)\propto i^{-2}$, in order
to measure how loose the bounds of \S\ref{sec:bound} are and to identify which
branch of Theorem~\ref{thm:floor} the recursion occupies. Throughout $m=40$,
$n=60$, $r=8$, $\beta_1=0.9$, and averages are over three seeds.

\paragraph*{Looseness.} Comparing each bound against the realised maximum over
$T=400$ steps at constant $\eta$:

\begin{center}
\begin{tabular}{lccc}
\toprule
$\gamma$ & $0.4$ & $0.7$ & $0.9$ \\
\midrule
Lemma~\ref{lem:Eeta}: bound / realised     & $3.2\times$ & $5.2\times$ & $10.4\times$ \\
Theorem~\ref{thm:bounded}: bound / realised & $30\times$  & --- & --- \\
\bottomrule
\end{tabular}
\end{center}

\noindent
Theorem~\ref{thm:bounded} applies only where $\kappa<1$, hence only in the first
column. Both bounds hold in every configuration tested; the slack in
Theorem~\ref{thm:bounded} is traced in
Remark~\ref{rem:notight} to the worst-case evaluation of
Lemma~\ref{lem:onestep} at a distortion ratio the dynamics need not visit.
Lemma~\ref{lem:Eeta} is the tighter of the two by roughly an order of magnitude,
which is consistent with its avoiding that step entirely.

\begin{figure}[h]\centering
\includegraphics[width=\linewidth]{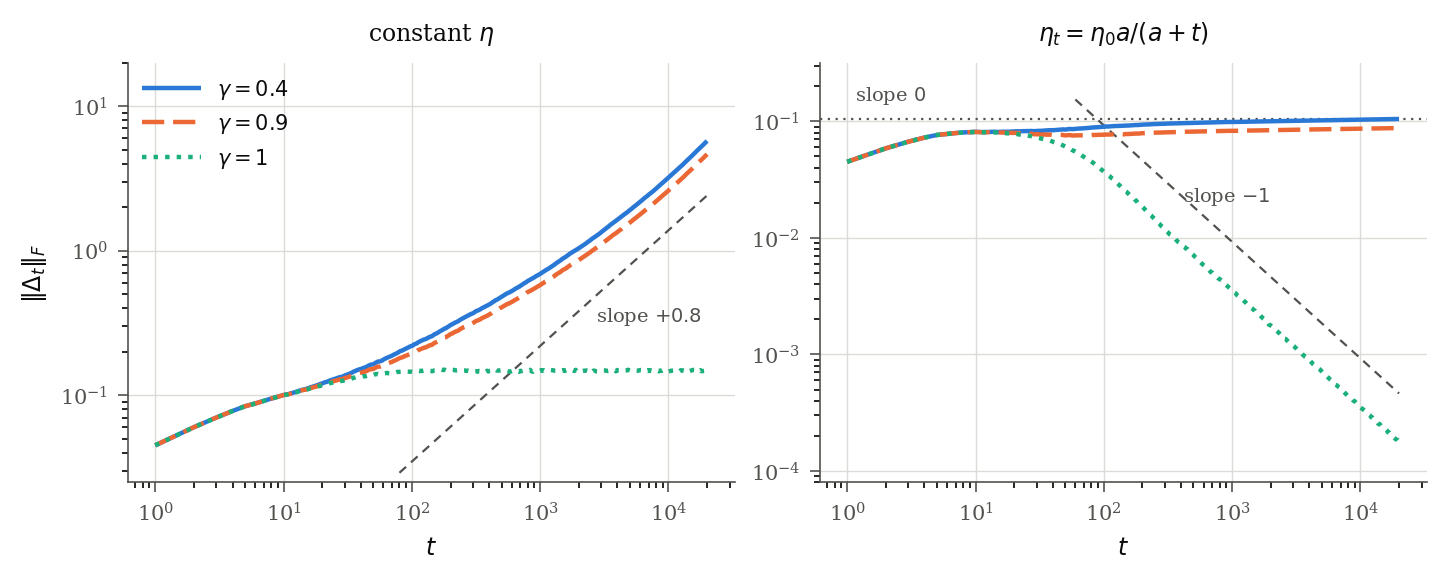}
\caption{Perturbed-iterate deviation $\Fnorm{\Delta_t}$ against $t$, log--log.
\emph{Left:} constant step size, where the deviation grows as $t^{0.83}$ for
every $\gamma<1$ --- an artefact of the schedule, not of the recursion.
\emph{Right:} the decaying schedule used by both reference analyses, where the
deviation is flat to the eye for $\gamma<1$ --- the floor branch of
Theorem~\ref{thm:floor}, with $D_T=O(\log^2T)$ by Corollary~\ref{cor:Dlog}
rather than the $\Theta(1)$ this panel alone would suggest; see
\S\ref{ssec:leaknum} and Figure~\ref{fig:leakage} --- and decays as $\eta_t$ for
$\gamma=1$ (the no-floor branch, $D=O(\eta_t^2)$). Dotted and dashed guides mark
slopes $0$ and $-1$. Averages over three seeds; identity is carried by dash pattern as
well as colour, so the panel is legible in greyscale.}\label{fig:deviation}
\end{figure}

Figure~\ref{fig:deviation} shows
$\Fnorm{\Delta_t}$ over $T=2\times10^4$ steps under the two step-size
conventions. The distinction that matters for Theorem~\ref{thm:floor} is not
between values of $\gamma$ but between schedules. At constant $\eta$ the
deviation grows as $t^{0.83}$ for every $\gamma<1$, placing the recursion in
neither branch; the mechanism is recorded in Remark~\ref{rem:constantstep}, and
the reason it is not the relevant regime is that both reference
analyses~\citep{stich2018,karimireddy2019} use decaying schedules. Under
$\eta_t=\eta_0a/(a+t)$ the growth all but disappears: the fitted exponents are
$+0.02$ at $\gamma=0.4$ and $\gamma=0.9$, and $-1.00$ at $\gamma=1$. The residual
$+0.02$ is not noise around zero. Theorem~\ref{thm:leak} predicts growth
proportional to $\sum_{u\le t}\eta_u=\Theta(\log t)$, which a power-law fit over
two decades reports as a small positive exponent, and
\S\ref{ssec:leaknum} separates the two readings by extending the horizon.

\noindent
That the $\gamma=1$ exponent is exactly $-1$ is the content of the second branch
of Theorem~\ref{thm:floor}: the deviation tracks the step size, so the
compression term enters the rate at order $\eta_t^3$ as in
\citet[Lemma~3.3]{stich2018} and no floor appears. For $\gamma<1$ the deviation
is flat to within a logarithm while $\eta_t\to0$ --- the ratio
$\Fnorm{\Delta_t}/\eta_t$ rises from $9.2$ at $t=10^2$ to $1751$ at
$t=2\times10^4$ --- which is the first branch. The next subsection measures the
leakage that produces it.

The constant-step column of Figure~\ref{fig:deviation} deserves its own record,
both because it is the second of the two negative results this paper reports and
because it is what justifies working under a decaying schedule throughout.

\begin{remark}[Constant step size is the wrong setting]\label{rem:constantstep}
Remark~\ref{rem:branch} reports growth exponents under a decaying schedule. At
\emph{constant} step size the picture is different and, for the present purpose,
misleading. Simulating \eqref{eq:recursion-eta} with persistent unit-norm
power-law gradients ($m=40$, $n=60$, $r=8$, $\beta_1=0.9$, $\eta\equiv0.05$,
$T=2\times10^4$, three seeds) and fitting on the last $60\%$:

\begin{center}
\begin{tabular}{lcccc}
\toprule
$\gamma$ & $0.4$ & $0.9$ & $0.99$ & $1.0$ \\
\midrule
exponent of $\Fnorm{S_t}$        & $0.83$ & $0.83$ & $0.79$ & $0.78$ \\
exponent of $\Fnorm{\Delta_t}$   & $0.83$ & $0.83$ & $0.77$ & $0.00$ \\
exponent of $\frac1T\sum_{t\le T}\Fnorm{\Delta_t}^2$
                                 & $1.56$ & $1.54$ & $1.40$ & $0.00$ \\
$\langle\tilde E_t,\tilde E_{t-1}\rangle_F/
 (\Fnorm{\tilde E_t}\Fnorm{\tilde E_{t-1}})$
                                 & $0.38$ & $0.85$ & $0.91$ & $0.91$ \\
\bottomrule
\end{tabular}
\end{center}

\noindent
The deviation grows like $t^{0.83}$ for every $\gamma<1$, which would place the
recursion in neither branch of Theorem~\ref{thm:floor}; the last row shows why,
the discarded tails being strongly correlated between consecutive steps, so that
$S_t$ accumulates a systematic direction rather than performing a random walk
(which would give exponent $0.5$). Both reference analyses use decaying
schedules, under which the growth disappears (Remark~\ref{rem:branch}), so the
constant-step regime is recorded only to justify that choice.
\end{remark}

\subsection{Measuring the leakage}\label{ssec:leaknum}

Figure~\ref{fig:deviation} could be read as showing the deviation
to be bounded under the decaying schedule, on the strength of a fitted exponent
of $+0.02$. Theorem~\ref{thm:leak} and Proposition~\ref{prop:lower} say what
that exponent is measuring: growth proportional to $\sum_{u\le t}\eta_u$, which
is logarithmic, and which over the two decades of a $2\times10^4$-step run is
almost indistinguishable from a constant. Three measurements separate the two
readings.

\paragraph*{The horizon test.} Extending the run of \S\ref{ssec:simulation} from
$T=2\times10^4$ to $T=5\times10^5$, three seeds, $\Fnorm{S_T}$ rises from
$0.01752$ to $0.01893$ at $\gamma=0.4$ and from $0.08745$ to $0.09401$ at
$\gamma=0.9$: a further $8.0\%$ and $7.5\%$ over a $25$-fold horizon. A
least-squares fit of $A+B\log t$ to the last $60\%$ of the \emph{short} run
($A=0.01380$, $B=3.8\times10^{-4}$ at $\gamma=0.4$) predicts $0.01881$ and
$0.09356$ at $T=5\times10^{5}$, within $0.8\%$ and $0.5\%$, whereas the reading
$\Fnorm{S_t}=\Theta(1)$ predicts the earlier values and is $8\%$ low. The
leakage is not bounded; it is logarithmic with a small coefficient.

\begin{figure}[h]\centering
\includegraphics[width=\linewidth]{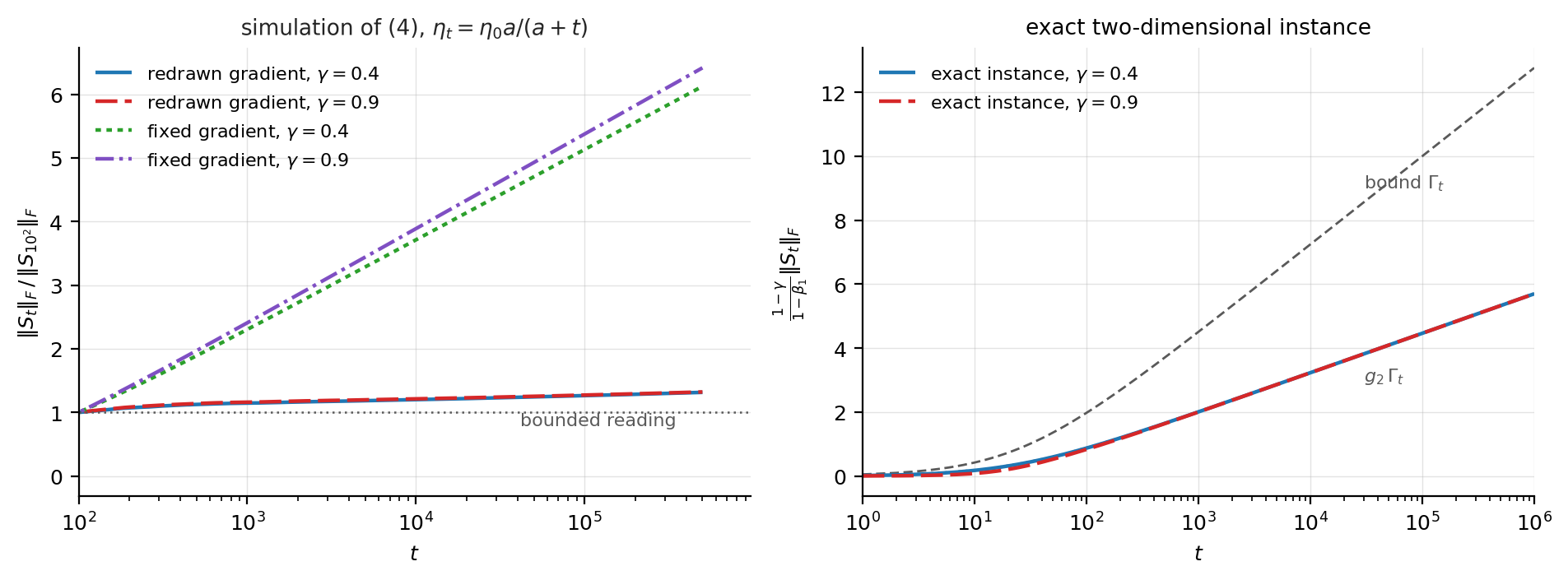}
\caption{The accumulated leakage grows logarithmically, and the factor
$1-\gamma$ cancels. \emph{Left:} $\Fnorm{S_t}$ from the simulation of
\eqref{eq:recursion-eta} under $\eta_t=\eta_0a/(a+t)$, normalised to its value
at $t=10^2$; a straight line on a logarithmic $t$-axis is growth proportional to
$\sum_{u\le t}\eta_u$, and the dotted line is the bounded reading. With the
gradient's singular subspaces redrawn at every step the coefficient is small and
two decades are not enough to see it; with a fixed gradient it is unmistakable.
\emph{Right:} the exact two-dimensional top-$1$ instance of
Proposition~\ref{prop:lower} ($c=0.5$), plotting the leakage contribution
$\frac{1-\gamma}{1-\beta_1}\Fnorm{S_t}$ to the deviation. The two values of
$\gamma$ give curves that agree to six decimal places --- the cancellation of
Theorem~\ref{thm:leak} --- and both track the lower bound $g_2\Gamma_t$ of
\eqref{eq:lower}, against the upper bound $\Gamma_t$ of
\eqref{eq:Sbound}.}\label{fig:leakage}
\end{figure}

\paragraph*{The persistent-gradient test.} The simulation of
\S\ref{ssec:simulation} redraws the gradient's singular subspaces at every step,
which is the most favourable case for cancellation. Repeating it with a fixed
gradient --- the situation of Proposition~\ref{prop:lower} --- makes the law
unmistakable. Over $T=5\times10^5$ steps the residual of the $A+B\log t$ fit is
$1.6\times10^{-4}$, against $2.8\times10^{-1}$ for a power law and $17.1$ for a
constant: the logarithm fits three orders of magnitude better than either
alternative. Fitting on the first $2\times10^4$ steps alone and extrapolating
predicts $\Fnorm{S_T}=0.04381$ at $T=5\times10^5$ against a realised $0.04409$,
an error of $0.6\%$, while $\Fnorm{S_T}$ itself has risen by $48\%$ over that
horizon. In the same runs
$\Fnorm{\Delta_T}=0.2645251212596051$ at $\gamma=0.4$ and $0.2645251212596045$
at $\gamma=0.9$ --- fourteen significant figures --- although $\Fnorm{S_T}$
differs between them by the factor $6=(1-0.4)/(1-0.9)$. This is the cancellation
of $1-\gamma$ in Theorem~\ref{thm:leak}, realised as an identity rather than as
an inequality.

\paragraph*{The exact instance.} The two-dimensional instance of
Proposition~\ref{prop:lower} ($\beta_1=0.9$, $\gamma=0.4$, $c=0.5$,
$\eta_t=0.05\cdot24/(24+t)$) run to $T=10^{6}$ gives $\Fnorm{S_T}=0.9496$
against the closed form $(1-\beta_1)g_2\Gamma_T/(1-\gamma)=0.9514$, and
increments per unit of $\log t$ of $0.0890$, $0.0897$ and $0.0879$ across the
decades $10^{3}$--$10^{6}$, against the predicted
$(1-\beta_1)g_2\eta_0a/(1-\gamma)=0.0894$. The growth is logarithmic to three
figures over three decades. The sufficient condition $c<1-\gamma\rho$ of
Proposition~\ref{prop:lower} is conservative: at $c=0.99$ the first coordinate
is still selected at every step for $\gamma$ up to $0.9$, and then
$\Fnorm{\Delta_T}/\Gamma_T=0.702$ against $g_2=0.7035$, so the constant in
\eqref{eq:Dbound} is attained to within $0.2\%$ of $1/\sqrt2$.

\paragraph*{The closed loop.} Corollary~\ref{cor:Dfinite} predicts that the
divergence disappears when the gradients decay along the trajectory. On
$f(x)=\frac12\sum_i\lambda_ix_i^2$ with top-$k$ compression this is what
happens: at $d=2$, $k=1$, $\lambda=(1,1)$ the leakage converges
($\Fnorm{S_t}=0.30398$ at $t=10^{5}$ and $0.30408$ at $t=2\times10^{5}$) and the
excess loss keeps falling, while at $d=100$, $k=10$ with $\lambda$ logarithmically
spaced over $[0.01,1]$ --- where most coordinates are never selected and the
gradient in them does not decay --- it grows logarithmically ($0.120$, $0.166$,
$0.198$, $0.207$ at $t=10^{3},10^{4},10^{5},2\times10^{5}$) and the excess loss
stalls near $2.9\times10^{-2}$ at $\gamma=0.4$ against $3.9\times10^{-3}$ at
$\gamma=1$. Both behaviours are the ones the corollary describes, and which of
them a given problem exhibits is decided by $\Gamma_\infty$, not by the
compressor.

\paragraph*{Bound checks.} Along the trajectories of \S\ref{ssec:simulation} at
$\gamma\in\{0,0.4,0.9,0.99\}$, under both schedules and both gradient models,
the maximum ratio of realised value to bound is $0.52$ for
\eqref{eq:Eeta}, $1.00$ for \eqref{eq:Bbound} (attained to four figures at a
constant gradient and constant step size), $0.085$ for the leakage bound in
\eqref{eq:Sbound} and $0.47$ for \eqref{eq:Dbound}; the splitting identity
$\Fnorm{\beta_1\tilde M+\gamma\tilde E}
=(\beta_1^2\Fnorm{\tilde M}^2+\gamma^2\Fnorm{\tilde E}^2)^{1/2}$ used in the
proof holds to $2.5\times10^{-15}$. No violation occurs.

The leakage bound is the loosest of the four, and the looseness is worth
decomposing because it is the whole of the remaining gap. Its coefficient of
$\log t$ is $(1-\beta_1)G\eta_0a/(1-\gamma)=0.2$ at $\gamma=0.4$; the realised
coefficient is $4.4\times10^{-3}$ with a fixed gradient and
$3.8\times10^{-4}$ with the gradient redrawn each step. The first factor,
$3.9$, is that the discarded tail is a fraction of the injected gradient
($\Fnorm{\tilde E_t}/\eta_t=0.044$ against the bound's $0.171$). The second,
$133$, is alignment: the normalised
$\langle S_{t-1},\tilde E_t\rangle_F$ settles at $+0.0075$ over the decades
$10^4$--$5\times10^5$ rather than at the $1$ the triangle inequality assumes.
It settles rather than decays, and $0.0075\times0.044\times\eta_0a
=4.0\times10^{-4}$ recovers the observed $3.8\times10^{-4}$ to $5\%$: the same
alignment that makes the bound loose is what makes the growth logarithmic
instead of bounded.

\subsection{Transformer measurement}\label{ssec:transformer}

We instrument \textsc{bert}-base~\citep{devlin2019} fine-tuning on
\textsc{glue}~\citep{wang2019}, running \eqref{eq:recursion} alongside the
optimizer and logging the singular values of $B_t$ and $G_t$ every $25$ steps
for six weight matrices spanning attention, \textsc{mlp} and depth. At $r=32$,
$p=768$, batch $16$, $600$ steps, the distortion ratio converges and

\begin{center}
\begin{tabular}{lccc}
\toprule
 & MRPC & CoLA & MRPC (depth-spread) \\
\midrule
$q$ (converged)                 & $0.295$ & $0.217$ & $0.259$ \\
$\alpha_T=1-q^2$                & $0.913$ & $0.953$ & $0.933$ \\
gain over $r/p=0.042$           & $21.9\times$ & $22.9\times$ & $22.4\times$ \\
gradient decay exponent         & $2.62$  & $4.58$  & $2.46$ \\
stable rank $B_t$ / $G_t$       & $3.7/3.1$ & $3.6/2.9$ & $3.5/2.9$ \\
\bottomrule
\end{tabular}
\end{center}

\noindent
Per-layer gains span $21.5$--$23.2\times$ across layer type and depth. The
\textsc{bert} pooler is excluded as degenerate (stable rank $1.16$--$1.30$: a
single dense map on the \texttt{[CLS]} token, whose gradient is near rank one).
Two observations bear on the theory. First, the gradient decay exponent varies
by nearly a factor of two between tasks while $q$ moves only from $0.22$ to
$0.30$, consistent with Theorem~\ref{thm:Ebound}, whose bound involves only
$\Fnorm{G_t}$ and not the gradient spectrum. Second, the $\gamma$ crossover
predicted by Theorem~\ref{thm:bounded} is observed: at $\gamma=1$
($\kappa=1.345$, no guarantee) $q$ climbs past $0.87$ without converging and the
stable rank of $B_t$ reaches $22.4$ against a gradient stable rank of $3.1$,
whereas at $\gamma=0.4$ ($\kappa=0.985$, guarantee applies) $q$ converges at
$0.26$ and the stable rank of $B_t$ is $3.5$. The low stable ranks are
consistent with the concentration of gradient energy in a small subspace
reported in~\citet{gurari2018,sagun2018}.

\emph{Limitations.} The two MRPC runs share a seed and differ only by the addition
of one layer, so there are two independent measurements rather than three;
additional seeds would strengthen the claim.

\subsection{Training quality at $\gamma<1$}\label{ssec:quality}

Proposition~\ref{prop:pi} identifies the deviation $\Delta_t$ exactly, for any
objective, and Theorem~\ref{thm:floor} shows that in the strongly convex setting
a deviation which does not shrink with the step size costs an additive floor.
Neither statement settles what $\gamma<1$ costs a transformer, so this
subsection measures it directly, and the remarks that follow instrument the
network to explain what is found.

We fine-tune \textsc{bert}-base on \textsc{mrpc} under
\eqref{eq:recursion-eta}, compressing the $72$ encoder attention and
\textsc{mlp} weight matrices at $r=32$ against $p=768$, so $r/p=0.042$. All
remaining tensors --- embeddings, \textsc{layernorm}, biases, the pooler and
the classification head --- run plain momentum \textsc{sgd} with the same
$\beta_1$ and $\eta$ in every arm, so the comparison isolates the compression.
The optimizer is the unpreconditioned recursion of
Assumption~\ref{as:alg}, hence plain \textsc{sgd}: $\eta=10^{-2}$ with
gradient clipping at $1$, chosen by a sweep with compression disabled, batch
$16$, two epochs ($458$ steps), $\beta_1=0.9$, three seeds, and an uncompressed
momentum-\textsc{sgd} control arm. Because the arms share a seed they share an
initialisation and a data order, so the comparison is paired and the
seed-to-seed variance cancels.

All three arms train. Against a majority-class baseline of $0.812$ in $F_1$ and
$0.684$ in accuracy, every arm reaches $F_1\approx0.843$ and accuracy
$\approx0.765$. The compression is active and not vacuous: the compressed arms
take $3.9$ minutes against $1.9$ for the control. The paired differences are

\begin{center}
\begin{tabular}{lccc}
\toprule
comparison & mean $\Delta F_1$ & $t(2)$ & $95\%$ CI \\
\midrule
$\gamma=1.0$ against $\gamma=0.4$        & $-0.00014$ & $-0.08$ & $[-0.0079,+0.0076]$ \\
$\gamma=1.0$ against uncompressed        & $+0.00014$ & $+0.05$ & $[-0.0123,+0.0126]$ \\
$\gamma=0.4$ against uncompressed        & $+0.00028$ & $+0.22$ & $[-0.0052,+0.0058]$ \\
\bottomrule
\end{tabular}
\end{center}

\noindent
Accuracy agrees to four decimal places across all three arms. The result is
therefore not merely an absence of evidence but a bound: on this task, any
effect of $\gamma$ on validation $F_1$ is smaller than $0.008$, under one
percentage point, and any effect of compression at all --- $\gamma=1$ against
no compression --- is smaller than $0.013$.

\begin{remark}[Why the floor is invisible here]\label{rem:notransfer}
The deviation of Proposition~\ref{prop:pi} is not small on this task, and
Theorem~\ref{thm:floor} attaches a cost of order $(\mu+2L)D$ to it wherever that
theorem applies; yet the null above shows none. Instrumenting the network
resolves the apparent tension.
Logging $\Fnorm{\Delta_t}$ during training, at $\eta=10^{-1}$ and one epoch,
gives at the final step
\begin{center}
\begin{tabular}{lcc}
\toprule
 & $\gamma=0.4$ & $\gamma=1$ \\
\midrule
$\Fnorm{X_0}$, pretrained weights          & $357.6$ & $357.6$ \\
$\Fnorm{X_T-X_0}$, fine-tuning displacement & $0.907$ & $0.951$ \\
$\Fnorm{\Delta_T}$                          & $0.422$ & $0.287$ \\
$\Fnorm{\Delta_T}/\Fnorm{X_T-X_0}$          & $0.465$ & $0.301$ \\
$\Fnorm{\Delta_T}/\Fnorm{X_0}$              & $0.118\%$ & $0.080\%$ \\
\bottomrule
\end{tabular}
\end{center}

\noindent
The deviation is roughly half the fine-tuning displacement but one part in a
thousand of the model, and the two ratios differ by a factor of about $400$.
That is the whole of the discrepancy. A problem started far from its solution
must travel the entire distance to $X^\star$, so a deviation worth half the
journey is worth half the solution and dominates the loss; fine-tuning starts at
a good point, moves $0.25\%$ of the weight norm, and the same
relative-to-journey deviation is a $0.1\%$ perturbation of a network that
tolerates far worse. The floor of Theorem~\ref{thm:floor} is a proved
consequence within its hypotheses; what the instrumentation shows is why no
trace of it appears at this scale of displacement.

We state the limitation this leaves. The argument turns on the displacement
being small relative to the weights, and it does not establish that the floor is
harmless where that is false --- pretraining from scratch being the obvious
case, and the one in which a large displacement is not a modelling artefact at
all.
\end{remark}

\begin{remark}[Two costs, not one]\label{rem:twocosts}
The same instrumentation records the distortion ratio and the accumulated
leakage, and these separate the two values of $\gamma$ in opposite directions.

\begin{center}
\begin{tabular}{lcc}
\toprule
over one epoch & $\gamma=0.4$ ($\kappa=0.985$) & $\gamma=1$ ($\kappa=1.345$) \\
\midrule
$q_t$, start to end            & $0.163\to0.202$, stable & $0.342\to0.763$, climbing \\
stable rank of $B_t$           & $2.6$--$5.2$ & $8.4$--$25.6$ \\
$\Fnorm{S_T}$                  & $0.065$ & $2.218$ \\
growth of $\Fnorm{S_t}$        & $139\times$ & $2863\times$ \\
\bottomrule
\end{tabular}
\end{center}

\noindent
Three readings follow. First, the crossover of \S\ref{ssec:transformer} is
reproduced with the compression driving training rather than observing it
alongside another optimizer, and it is sharper here: at $\gamma=1$, where
$\kappa>1$ and Theorem~\ref{thm:bounded} gives no guarantee, $q$ does not
converge and the accumulator's stable rank reaches $25.6$ against a gradient
stable rank near $3$, so the truncation is discarding most of what it is given.
At $\gamma=0.4$, where the guarantee applies, both are stable.

Second, Proposition~\ref{prop:pi} is confirmed on a real network in the one
place it makes a counter-intuitive prediction. The accumulated leakage
$\Fnorm{S_T}$ is $34$ times larger at $\gamma=1$ and grows by three orders of
magnitude during training, yet $\Fnorm{\Delta_T}$ is \emph{smaller} there,
because the coefficient $(1-\gamma)$ annihilates $S_t$ exactly. The buffer
diverges and the deviation does not.

Third, and consequently, $\gamma=1$ is not simply the benign end of the range.
It removes the leakage term from \eqref{eq:pi}, and it forfeits the boundedness
of the accumulator. Below $1$ the cost is a bias; at $1$ it is a compression
that grows steadily lossier. The two are visible in the same run, and neither
dominates on this task --- which is consistent with the null above and suggests
that an intermediate $\gamma$, rather than either endpoint, is what one should
expect to be optimal.
\end{remark}

The sweep above is at $\eta=10^{-2}$ alone, so a second operating point is
worth recording. Repeating it at $\eta=10^{-1}$, the value the step-size sweep rates highest, with one epoch and five seeds, raises every arm --- uncompressed to $0.882$, $\gamma=0.4$ to $0.880$ and $\gamma=1$ to $0.871$ in $F_1$ --- while leaving the comparison null. Against $\gamma=0.4$, $\gamma=1$ gives $-0.0096$ with $t(4)=-0.70$ and a $95\%$ interval of $[-0.048,+0.028]$. That interval is markedly \emph{wider} than at $\eta=10^{-2}$, the paired standard deviation having risen from $0.0031$ to $0.0306$; the better-performing configuration is the less sensitive experiment, and five seeds there resolve less than three do at the lower step size. We therefore report $\eta=10^{-2}$ as the primary evidence and $\eta=10^{-1}$ as corroboration. The point estimates at $\eta=10^{-1}$ place $\gamma=1$ last rather than first, the direction Remark~\ref{rem:twocosts} would lead one to expect, but the effect is well inside the noise and we draw nothing from it.

\section{Related work}\label{sec:related}

Retaining what a compressor discards and re-injecting it at the next step
originates with 1-bit \textsc{sgd}~\citep{seide2014}. The first convergence
analyses are due to \citet{stich2018} in the convex
setting and \citet{karimireddy2019} in
the non-convex setting; both proceed through a perturbed-iterate argument and
both bound the memory through a compressor contraction constant, which is the
step \S\ref{ssec:stage2} replaces. \citet{stich2020}
later unified the two settings and obtained a rate splitting into a stochastic
term unaffected by compression and a higher-order term only linearly slowed.
That separation is the scalar precedent for the phenomenon established here
for matrix-valued momentum, and Theorem~\ref{thm:floor} may be read as
identifying what replaces the higher-order term when the deviation fails to
shrink with the step size.

The recursion analysed in this paper is that of
\textsc{umtam}~\citep{moayedikia2026}, which introduced it as the training core
of a unified training-and-merging framework and whose theoretical analysis
concerns the full preconditioned method --- convergence rates, merging
quality, and generalization bounds. On the buffer itself that paper records a
steady-state estimate, $\Fnorm{E_\infty}\le\bar\sigma_{r+1}/(1-\gamma)$
\citep[Lemma~4]{moayedikia2026}, obtained by treating the per-step truncation
error as stationary. Theorem~\ref{thm:Ebound} replaces that estimate with an
unconditional bound along the trajectory, proved rather than posited, and
nothing else in \S\ref{sec:bound}--\S\ref{sec:numerics} --- the competitor
argument, the dimension-free Stage-2 constant, the leakage term and its lower
bound, or the measurements --- appears there. The present paper is the
companion analysis of that recursion, and should be read alongside it.

EF21~\citep{richtarik2021} removed the bounded-gradient assumption and
introduced the biased-compressor class $\mathcal B(\alpha)$ used in
Corollary~\ref{cor:class}; a subsequent refinement~\citep{ef2024} sharpens the
EF21 rate from the quadratic to the arithmetic mean of the per-node smoothness
constants, within that same class --- which explicitly names the rank-$r$
approximator as a member. The complementary case of unbiased quantization, for
which no memory is required, is treated by \citet{alistarh2017}. Error feedback has also been analysed in the
federated setting with partial participation and an adaptive base
method~\citep{fedef2023}.

The closest reference point for the present work is the analysis of error
feedback applied to a momentum-like quantity. \citet{fatkhullin2023} show that momentum improves error-feedback
rates, and any claim about the pairing must be measured against theirs; a
recent line removes the need to know smoothness constants when tuning the step
size in the same setting~\citep{paramagnostic2025}, and error feedback has been
carried beyond the Euclidean geometry altogether by an \textsc{lmo}-based
construction~\citep{muon2025}, which is the nearest neighbour in setting to the
matrix-valued recursion studied here; both are read against
Theorem~\ref{thm:Ebound} at the end of this section. The contribution here is
not the pairing but the observation that when the \emph{compressed} momentum
is fed back before truncation, that compressed state is itself an admissible
competitor in the Eckart--Young minimisation (\S\ref{ssec:necessity}). This is
a structural property of the recursion rather than a rate improvement obtained
from a better choice of momentum parameters.

Low-rank compression of gradients and of optimizer state has developed largely
in parallel. PowerSGD~\citep{vogels2019} established the approach as a
practical distributed method, assuming rather than proving that the
best-rank-$r$ map is contractive; Corollary~\ref{cor:class} supplies that
statement, and GreedyLore~\citep{chen2025} independently establishes
contractivity in the semi-lazy regime together with a linear-speedup rate,
while a further refinement~\citep{powersgdplus2025} formalises the optimal
rank-$r$ compressor inside an error-feedback proof and, more pointedly,
exhibits a counterexample on which PowerSGD fails to converge --- direct
evidence that the contractivity assumed in~\citet[Assumption~A]{vogels2019}
needed establishing rather than positing.

GaLore~\citep{zhao2024} projects gradients into a low-rank subspace to compress
optimizer state, and LDAdam~\citep{robert2025} adds projection-aware updates
and a generalized error-feedback mechanism covering both gradient and state
compression; both invoke a contraction whose constant is $\Theta(1-r/p)$ in
the worst case. A complementary remedy for the bias of deterministic low-rank
projection, using randomisation rather than feedback, is given
in~\citet{golore2025}, and cheaper alternatives that approximate the
truncated-SVD step by alternating least squares or subspace iteration have
been proposed by a single line of work~\citep{oplora2025,psilora2026}, the
earlier of the two in workshop form; the exact projection is retained here
because the competitor argument of \S\ref{ssec:necessity} depends on it.

Adafactor~\citep{shazeer2018} supplies the rank-one row/column factorization of
the second moment used by the full \textsc{umtam}
algorithm~\citep{moayedikia2026}, which is why the preconditioner is set aside
in Assumption~\ref{as:alg} rather than absorbed into it.
Architecturally the closest neighbour is EFCP~\citep{modoranu2024}, which
compresses the gradient window --- by sparsification or by low rank ---
\emph{before} it enters the preconditioner and feeds the residual back. EFCP
does not bound a rank-free momentum buffer, so Theorem~\ref{thm:Ebound} is not
subsumed by it, but the compress-before-preconditioner architecture is theirs
and the present contribution rests on the narrower ground of the
constant and its proof device.

The same group's MicroAdam~\citep{modoranu2024micro} is closer still in one
respect and further in another: it compresses the gradient by top-$k$
sparsification before the information enters the optimizer state, controls the
resulting error by feedback, and --- its stated innovation over EFCP, which
``does not compress the error accumulator'' \citep[\S2]{modoranu2024micro} ---
compresses the feedback buffer itself by quantization. The compressed object
there is still the gradient, and the buffer is quantized rather than
re-injected before a projection, so the competitor of \S\ref{ssec:necessity}
is again unavailable; but MicroAdam is the nearest single-node
error-feedback-in-the-optimizer-state work and belongs in any comparison.

APOLLO~\citep{zhu2025} sits in the GaLore lineage and is included because it
makes the opposite architectural choice, avoiding the truncated-\textsc{svd}
step entirely in favour of a channel-wise scaling rule, which is the natural
baseline for the cost objection to $\Pr_r$. Shampoo~\citep{gupta2018} and
SOAP~\citep{vyas2024} represent the structured full-matrix preconditioning
line, which compresses curvature rather than the update and is therefore
complementary; the reading of a diagonal adaptive preconditioner as a
natural-gradient metric~\citep{fadam2024} would be the natural framework in
which to absorb the row/column statistics, and the general equivalence between
preconditioned steps and Riemannian gradient descent is recorded
in~\citet{rlo2026}.

The compression map $\Pr_r$ is the metric projection onto the determinantal
variety, and its well-posedness away from singular-value crossings is due to
\citet{absil2012}. The manifold $\mathcal M_r$, its tangent
space and its retractions are developed by
\citet{vandereycken2013} within the framework of~\citet{absil2008};
perturbation of the rank-$r$ approximation itself, rather than of its singular
subspaces, has been bounded in terms of the relative spectral
gap~\citep{relgap2023}, and Lipschitz estimates for functions of singular
values are classical~\citep{oplip2015}.

\citet{mishra2016} study metrics on low-rank quotient
manifolds derived from second-order information, and \citet{tong2021} precondition the \emph{factors} of a low-rank iterate with
$r\times r$ matrices to obtain condition-number-free convergence, an approach
later extended to overparameterised factors~\citep{scaledgdover2023}; both are
adjacent to the row/column preconditioner of the full algorithm, which acts
elementwise on the matrix rather than on its factors. A geometric treatment of
momentum-based optimizers constrained to low rank has also been
proposed~\citep{geomlowrank2025}.

The dynamical low-rank literature has treated the behaviour of low-rank
iterates near the manifold boundary in considerably more depth than the
optimization literature: \citet{koch2007} introduce the
setting, \citet{kieri2016} analyse small singular
values, \citet{ceruti2022a} and \citet{ceruti2022b} give robust and rank-adaptive integrators, and
retractions approximating the truncated SVD have been shown to converge
geometrically to the best low-rank approximation~\citep{charous2022}. These
address a matrix ODE rather than an optimizer state, but any treatment of
adaptive rank in the present setting should be positioned against them; on the
machine-learning side, AdaLoRA~\citep{zhang2023} adjusts rank online in an SVD
parameterisation by pruning singular values. The Hadamard-product rank
identities relevant to the omitted preconditioner are classical, and the sharp
lower bound is due to \citet{horn2020}.

A separate line studies precisely the operator this paper compresses with, and
it sets the standard of care that \S\ref{ssec:recursion} and
Remark~\ref{rem:selection} try to meet. \citet{olikier2024lowrank} treat $\Pr_r$ as the set-valued map it is ---
$\Pr_r(B)$ is the set of all rank-$\le r$ truncated \textsc{svd}s of $B$,
nonempty and compact --- and every algorithm in that literature is written
with an arbitrary selection, every theorem quantified over all selections,
with no gap condition anywhere.

Two further findings from it bear on any use of $\Pr_r$ inside an iteration.
First, $\R^{m\times n}_{\le r}$ fails to be Clarke regular at every point of
its singular part $\{\rank<r\}$, so Bouligand stationarity is there strictly
stronger than the Mordukhovich notion, and the usual stationarity measure can
fail to be lower semicontinuous exactly there; \citet{olikier2024pgd} exhibit a first-order method on
$\R^{3\times3}_{\le2}$ converging to a rank-one point that is Mordukhovich but
not Bouligand stationary, with the measure tending to zero throughout, and
prove (their Theorem~1.2) that projected gradient descent with a nonmonotone
backtracking line search does accumulate at Bouligand stationary points on an
arbitrary closed set.

Second, they record as \emph{open} whether that guarantee extends to search
directions other than the negative gradient. A compressed-momentum direction
is exactly such a direction, so the present recursion sits on the open side of
that question; nothing here should be read as a stationarity claim on the
variety, and \S\ref{sec:conseq} makes none --- the guarantees are for the
parameter iterate $X_t$ in the ambient space, with $\Pr_r$ acting on the
\emph{update}, not on the iterate.

Two cost observations complete the picture. \citet{absil2015} survey retractions on $\mathcal M_r$, confirm that
the truncated \textsc{svd} is the metric projection and a bounded second-order
retraction, and note that when the argument is a tangent update the projection
needs only a $2r\times2r$ \textsc{svd} at $O((m+n)r^2)$ flops --- which does
\emph{not} apply here, since $\gamma E_{t-1}$ keeps $B_t$ dense. \citet{shalit2012} take the opposite route for online
low-rank learning, replacing the projection by a cheaper second-order
retraction on the ground that repeated \textsc{svd} is prohibitive; they
maintain no residual and so have no truncation error to recycle, which makes
them the natural foil for the present design.

A pattern runs through the low-rank branch of this literature that is worth
stating explicitly, since it is what the present paper acts on. Two
independent lines --- PowerSGD~\citep[Assumption~A]{vogels2019} and
LDAdam~\citep{robert2025} --- posit a contraction for best-rank-$r$ compression
and justify it by appeal to the empirically top-heavy spectrum of stochastic
gradients~\citep{gurari2018,sagun2018}. Corollary~\ref{cor:class} shows the
assumption holds with $\alpha=r/p$, and that this constant is tight over all
inputs. Section~\ref{ssec:wrong} then argues that tightness over all inputs is
the wrong criterion, since the flat spectra attaining it are not visited by
the dynamics, and the measurements of \S\ref{ssec:transformer} quantify the
gap at a factor of $22$.

Two neighbouring results deserve an explicit answer, since each might be read
as bearing on the present one.

GreedyLore~\citep{chen2025} separates a \emph{lazy} from a \emph{semi-lazy}
update of the compression basis: the lazy variant computes a truncated
\textsc{svd} once every $\tau$ iterations and holds the resulting projector
fixed in between, while the semi-lazy variant they propose refreshes the
selected columns at every step between full recomputations. For the lazy
variant they prove non-contractivity outright --- their Proposition~2 exhibits
a gradient on which the compression error equals the full gradient norm, which
is what a projector frozen onto a stale subspace does to a gradient that has
since rotated out of it.

The objection does not reach the recursion studied here.
Assumption~\ref{as:alg} recomputes $\Pr_r(\tilde B_t)$ at every step, from the
current accumulator and with no carried-over basis, so the compressor is at
each step a genuine metric projection onto $\R^{m\times n}_{\le r}$ and the
counterexample has nothing to freeze. The independence is in fact stronger
than this comparison suggests: contractivity is a hypothesis
Theorem~\ref{thm:Ebound} never invokes, so even a compressor to which their
Proposition~2 applied would not disturb it, provided the compressed momentum
remains an admissible competitor in the minimisation. What their result does
establish is the sharpness of the distinction drawn in \S\ref{ssec:relation}:
a stale basis and a recomputed one are different compressors, and only the
latter is the metric projection this paper's argument uses.

The low-rank Adam of~\citet{geomlowrank2025} is a different mechanism from the
row/column preconditioner omitted in Assumption~\ref{as:alg}, and does not
supply its rank behaviour. Their second moment is carried as a full $r\times
r$ coefficient matrix in the shared bases of the low-rank factorisation,
projected through an elementwise square root and re-squared to keep it
positive; rank is then \emph{adapted}, by augmenting the bases to span both
the current iterate and the gradient --- which doubles the rank --- and
truncating back, either to $r$ or at a relative threshold.

The omitted preconditioner does the opposite on both counts. It is the
rank-one row/column factorization of Adafactor~\citep{shazeer2018}, acting
elementwise on the full $m\times n$ matrix, and a Hadamard product with a
rank-one matrix of nonzero entries leaves the rank of the compressed momentum
exactly where it was~\citep{horn2020}. So the two are complementary rather than
overlapping: \citet{geomlowrank2025} changes the rank of the iterate and
preconditions inside the coefficient space, whereas the preconditioner set
aside here preserves rank and acts outside it. The question of what that
preconditioner does to the results of \S\ref{sec:bound} is therefore still
open, and is not answered by the prior work.

The closest neighbours in the error-feedback literature are two recent papers
that compress a \emph{momentum difference}, which is nearer to the present
recursion than the other entries of Table~\ref{tab:schemes}, and they deserve
an explicit answer. EF21-Muon~\citep{muon2025} extends
error feedback beyond the Euclidean setting: each worker forms a momentum
$M^{k+1}_j$, transmits $R^{k+1}_j=\mathcal C^k_j(M^{k+1}_j-G^k_j)$ and updates
its state by $G^{k+1}_j=G^k_j+R^{k+1}_j$. The parameter-agnostic analysis
of~\citet{paramagnostic2025} has the same shape across five momentum schemes,
compressing $v^{t+1}_i-g^t_i$ and updating $g^{t+1}_i=g^t_i+c^{t+1}_i$.

In both, the compressed state enters the argument of the next compression
\emph{with a minus sign} and, being an accumulation of compressed increments,
need not lie in the compressor's range at all; neither therefore supplies the
admissible competitor that Theorem~\ref{thm:Ebound} runs on, and both retain
the contraction constant that it removes. The former assumes $\E\lVert\mathcal
C(X)-X\rVert^2\le(1-\alpha)\lVert X\rVert^2$ and carries $(1-\alpha)/\alpha$
and $\alpha^{-2}$ into its stochastic rate; the latter carries
$\sqrt{1-\alpha}$ in its tracking lemma and $\alpha^{-2}$ into each of its
five theorems.

A second difference bears on \S\ref{sec:split} rather than \S\ref{sec:bound}:
the oracle of~\citet{muon2025} is a linear minimisation oracle over a norm
ball, $\min_{Z\in\mathcal B(X,t)}\langle G,Z\rangle$, and not a metric
projection onto a set, so the orthogonal splitting of Lemma~\ref{lem:split}
has no counterpart there and is not needed. The same reading applies to the
tight analyses of \textsc{ef} and \textsc{ef21} obtained through performance
estimation by \citet{tightef2026}: what is
tight there is the rate \emph{given} a contraction constant, which is assumed
of the compressor and appears in every bound, whereas the question asked here
is whether that constant need enter the buffer bound at all.

\section{Conclusion}\label{sec:concl}

Compressing the momentum, and re-injecting the compressed momentum before the
next truncation, removes the compressor's contraction constant from the buffer
bound entirely. The proof turns on a single observation: the
quantity being bounded is a distance to a cone, and the algorithm's own state
already lies in that cone. Because nothing else about the compressor is used,
the statement is not about truncated \textsc{svd}; it holds for top-$k$
selection and for any other metric projection onto a closed cone, and it holds
for every selection from the projection set, so no condition on singular-value
crossings is needed. In its trajectory-tracking form it needs no assumption on
the gradients either.

The consequence for rates is partial and we have tried to be exact about where
the line falls. Stage 2 of the standard template closes with a
constant free of $m$, $n$ and $r$. Stage 1 does not: the perturbed-iterate
deviation carries an accumulated-leakage term with no counterpart in the
reference analyses, and Theorem~\ref{thm:floor} shows that if this term is
merely bounded, rather than shrinking with the step size, the price is an
additive floor and not a slower rate. Theorem~\ref{thm:leak} supplies the
missing bound: the leakage never exceeds the step-weighted gradient mass
$\Gamma_t$, with the factor $1-\gamma$ cancelling exactly, so the hypothesis of
Theorem~\ref{thm:floor} holds unconditionally with $D_T=O(\log^2T)$ and holds
with a constant whenever $\Gamma_\infty$ converges. It does not hold uniformly:
Proposition~\ref{prop:lower} exhibits a constant gradient and a top-$1$
compressor for which the deviation diverges logarithmically, which is why a
fitted growth exponent of $+0.02$ over $2\times10^4$ steps, easily read as a
bounded deviation, is in fact measuring a logarithm.
What remains open is the gap between the two: with the discarded tails
unaligned the realised leakage grows about five hundred times more slowly than
the bound, and describing that regime needs the alignment
$\langle S_{t-1},\tilde E_t\rangle_F$ rather than the norms.

Several limitations should be read alongside the results. The bounds are loose
by factors of three to thirty against realised trajectories
(\S\ref{ssec:tightness}), and the behaviour as $\gamma\uparrow1$ is covered only
by a sufficient condition that simulation shows to be pessimistic. The
dimension-free \emph{buffer} bound needs no bounded-gradient assumption, but the
dimension-free \emph{Stage-2 constant} still does, and removing it would require
tracking the discounted gradient sum inside a Lyapunov argument
(Remarks~\ref{rem:bgfree} and~\ref{rem:bg}). The analysis is single-node and omits the row/column
preconditioner of the full \textsc{umtam} algorithm~\citep{moayedikia2026}. It
makes no claim about stationarity on
the determinantal variety, where the relevant notion is subtle and the available
guarantees are for projected gradient descent rather than for a momentum
direction.

The empirical picture closes with a reconciliation and one limitation. At
$r/p=32/768$ --- compression as aggressive as a transformer uses --- fine-tuning
\textsc{bert} shows no cost from the feedback decay at all, and instrumenting
the network says why: the deviation is about half the fine-tuning displacement
but a thousandth of the pretrained weights, and it is the latter ratio that
quality responds to. The floor of Theorem~\ref{thm:floor} is a proved
consequence; what the measurement adds is that it is invisible at this scale of
displacement. The limitation is that this argument turns on
fine-tuning moving a short distance from a good initialisation, so it says
nothing about training from scratch, where a problem started far from its
solution is not a modelling artefact. The same instrumentation also shows that
$\gamma=1$ is not the benign endpoint it appears to be in the analysis: it
removes the leakage term and forfeits boundedness of the accumulator, whose
distortion ratio climbs to $0.76$ and whose stable rank grows by an order of
magnitude. There are two costs, not one, and the interesting value of $\gamma$
is likely to lie between them.

\paragraph*{Code availability.} The code for the simulations and
instrumentation of \S\ref{sec:numerics} is available in the repository of the
companion paper~\citep{moayedikia2026},
\url{https://github.com/amoayedikia/UMTAM}.

\vskip 0.2in
\bibliography{bib/references}

\appendix

\section{Proofs}\label{app:proofs}

The short proofs are given with their statements in the body; collected here are
those long enough to interrupt it, in order of appearance. All statements have
been verified numerically.

\subsection*{Proof of Lemma~\ref{lem:split}}

\emph{The cone statement \eqref{eq:split}.} If $M=0$ then $E=B$ and
$\langle M,E\rangle_F=0$ trivially, so assume $M\neq0$ and set
\[
  g(\lambda)\;:=\;\Fnorm{B-\lambda M}^2
  \;=\;\Fnorm{B}^2-2\lambda\langle B,M\rangle_F+\lambda^2\Fnorm{M}^2 ,
  \qquad \lambda\ge0 .
\]
Because $S$ is a cone and $M\in S$, every $\lambda M$ with $\lambda\ge0$ lies in
$S$, so $g(\lambda)\ge\Fnorm{B-M}^2=g(1)$ for all $\lambda\ge0$ by the defining
property of $\Pr_S$ in Definition~\ref{def:cone}. Thus $g$, which is
differentiable, attains its minimum over $[0,\infty)$ at $\lambda=1$, an interior
point of that set, whence $g'(1)=0$. Since
$g'(\lambda)=-2\langle B,M\rangle_F+2\lambda\Fnorm{M}^2$, this reads
$\langle B,M\rangle_F=\Fnorm{M}^2$, i.e.\ $\langle M,B-M\rangle_F=0$, which is
$\langle M,E\rangle_F=0$. Expanding $\Fnorm{B}^2=\Fnorm{M+E}^2$ and using the
vanishing cross term gives the Pythagorean identity, and the expressions for
$\Fnorm{E}$ and $\Fnorm{M}$ in terms of $q(B)$ follow on dividing by
$\Fnorm{B}^2$. No property of $S$ beyond closure under nonnegative scaling has
been used.

\emph{The spectral statement \eqref{eq:qexact}.} Let now
$S=\R^{m\times n}_{\le r}$ and write $B=\sum_{i\le p}\sigma_iu_iv_i^{\top}$ with
$\{u_i\}$ and $\{v_i\}$ orthonormal. By \eqref{eq:proj},
$M=\sum_{i\le r}\sigma_iu_iv_i^{\top}$ and $E=\sum_{i>r}\sigma_iu_iv_i^{\top}$:
the two sums use disjoint singular triples. Since
\[
  \langle u_iv_i^{\top},u_jv_j^{\top}\rangle_F
  =\operatorname{tr}(v_iu_i^{\top}u_jv_j^{\top})
  =(u_i^{\top}u_j)(v_j^{\top}v_i)=\delta_{ij},
\]
orthonormality gives $\Fnorm{E}^2=\sum_{i>r}\sigma_i^2$, and dividing by
$\Fnorm{B}^2=\sum_i\sigma_i^2$ gives \eqref{eq:qexact}. (This route also
recovers $\langle M,E\rangle_F=\sum_{i\le r}\sum_{j>r}\sigma_i\sigma_j
\delta_{ij}=0$ directly, without the cone argument.) \qed

\subsection*{Proof of Proposition~\ref{prop:pi}}

Write $\Sigma_M:=\sum_{s=1}^{t}\tilde M_s$. Summing $X_{s+1}=X_s-\tilde M_s$
gives $X_{t+1}=X_1-\Sigma_M$. From \eqref{eq:recursion-eta},
\[
  \tilde M_s=\beta_1\tilde M_{s-1}+\eta_s(1-\beta_1)G_s+\gamma\tilde E_{s-1}
  -\tilde E_s .
\]
Summing over $s=1,\dots,t$ and using $\tilde M_0=\tilde E_0=0$, so that
$\sum_{s=1}^{t}\tilde M_{s-1}=\Sigma_M-\tilde M_t$ and
$\sum_{s=1}^{t}\tilde E_{s-1}=S_t-\tilde E_t$,
\[
  \Sigma_M=\beta_1(\Sigma_M-\tilde M_t)+(1-\beta_1)\sum_{s=1}^t\eta_sG_s
  +\gamma(S_t-\tilde E_t)-S_t .
\]
Collecting $\Sigma_M$ and dividing by $1-\beta_1>0$ gives
$\Sigma_M=\sum_s\eta_sG_s-\Delta_t$, which is \eqref{eq:pi}. The equivalent
formulation follows by differencing: $\hat X_{t+1}-\hat X_t
=(X_{t+1}-\Delta_t)-(X_t-\Delta_{t-1})=-\tilde M_t-\Delta_t+\Delta_{t-1}
=-\eta_tG_t$. \qed

\subsection*{Proof of Lemma~\ref{lem:Beta}}

By Lemma~\ref{lem:split} applied to $(\tilde B_s,\tilde M_s,\tilde E_s)$,
$\Fnorm{\tilde M_s}^2=\Fnorm{\tilde B_s}^2-\Fnorm{\tilde E_s}^2
\le\Fnorm{\tilde B_s}^2$. Hence, from \eqref{eq:recursion-eta} and the triangle
inequality,
$\Fnorm{\tilde B_t}\le\beta_1\Fnorm{\tilde B_{t-1}}
+(1-\beta_1)\eta_t\Fnorm{G_t}+\gamma\Fnorm{\tilde E_{t-1}}$, and unrolling from
$\tilde B_0=0$,
\[
  \Fnorm{\tilde B_t}\;\le\;(1-\beta_1)\sum_{i=0}^{t-1}\beta_1^{i}\eta_{t-i}
  \Fnorm{G_{t-i}}\;+\;\gamma\sum_{i=0}^{t-2}\beta_1^{i}\Fnorm{\tilde E_{t-1-i}} .
\]
Assumption~\ref{as:G} and $\eta_{t-i}\le\rho^{i}\eta_t$ bound the first sum by
$(1-\beta_1)G\eta_t/(1-\beta_1\rho)$. For the second, \eqref{eq:Eeta} gives
$\Fnorm{\tilde E_{t-1-i}}\le(1-\beta_1)G\eta_{t-1-i}/(1-\gamma\rho)
\le(1-\beta_1)G\rho^{i+1}\eta_t/(1-\gamma\rho)$, so it is at most
$\gamma\rho(1-\beta_1)G\eta_t/\big((1-\gamma\rho)(1-\beta_1\rho)\big)$. Adding,
\[
  \Fnorm{\tilde B_t}\;\le\;(1-\beta_1)G\eta_t
  \frac{(1-\gamma\rho)+\gamma\rho}{(1-\beta_1\rho)(1-\gamma\rho)}
  \;=\;\frac{(1-\beta_1)G\eta_t}{(1-\beta_1\rho)(1-\gamma\rho)} .
\]
The bounds for $\Fnorm{\tilde M_t}$ and $\Fnorm{\tilde E_t}$ follow from
$\Fnorm{\tilde M_t},\Fnorm{\tilde E_t}\le\Fnorm{\tilde B_t}$, again by
Lemma~\ref{lem:split}. \qed

\subsection*{Proof of Theorem~\ref{thm:leak}}

By Theorem~\ref{thm:Ebound} in the step-size-carrying form
\eqref{eq:Eeta-traj},
$\Fnorm{\tilde E_s}\le(1-\beta_1)\sum_{j=0}^{s-1}\gamma^{j}\eta_{s-j}
\Fnorm{G_{s-j}}$. Summing over $s\le t$ and exchanging the order of summation,
each index $u=s-j$ appearing once for each $s\ge u$,
\[
  \Fnorm{S_t}\;\le\;\sum_{s=1}^{t}\Fnorm{\tilde E_s}
  \;\le\;(1-\beta_1)\sum_{u=1}^{t}\eta_u\Fnorm{G_u}
  \sum_{j=0}^{t-u}\gamma^{j}
  \;\le\;\frac{1-\beta_1}{1-\gamma}\,\Gamma_t ,
\]
which is the first part of \eqref{eq:Sbound}. For the second, Lemma~\ref{lem:split}
gives $\langle\tilde M_t,\tilde E_t\rangle_F=0$, hence
\[
  \Fnorm{\beta_1\tilde M_t+\gamma\tilde E_t}
  =\big(\beta_1^2\Fnorm{\tilde M_t}^2+\gamma^2\Fnorm{\tilde E_t}^2\big)^{1/2}
  \le\kappa\big(\Fnorm{\tilde M_t}^2+\Fnorm{\tilde E_t}^2\big)^{1/2}
  =\kappa\Fnorm{\tilde B_t},
\]
using the Pythagorean identity \eqref{eq:split} in the last step. Applying the
triangle inequality to \eqref{eq:pi},
\[
  \Fnorm{\Delta_t}\le\frac{\Fnorm{\beta_1\tilde M_t+\gamma\tilde E_t}
  +(1-\gamma)\Fnorm{S_t}}{1-\beta_1}
  \le\frac{\kappa\Fnorm{\tilde B_t}}{1-\beta_1}
  +\frac{(1-\gamma)}{1-\beta_1}\cdot\frac{(1-\beta_1)\Gamma_t}{1-\gamma}
  =\frac{\kappa\Fnorm{\tilde B_t}}{1-\beta_1}+\Gamma_t .
\]
The factor $1-\gamma$ cancels exactly. Neither step uses
Assumption~\ref{as:G}, the schedule, or any property of $S$ beyond
Definition~\ref{def:cone}. Substituting \eqref{eq:Bbound} gives
\eqref{eq:Dbound}. \qed

\subsection*{Proof of Corollaries~\ref{cor:Dlog} and~\ref{cor:Dfinite}}

For $\eta_t=\eta_0a/(a+t)$ one has $\rho=(a+1)/a$, and $\beta_1\rho<1$,
$\gamma\rho<1$ hold under the stated condition on $a$ exactly as in
Corollary~\ref{cor:shift}. Under Assumption~\ref{as:G},
$\Gamma_t\le G\sum_{u=1}^{t}\eta_u
\le G\eta_0a\int_0^{t}\frac{du}{a+u}=G\eta_0a\log(1+t/a)$, and $\eta_t\le\eta_0$,
so \eqref{eq:Dbound} gives \eqref{eq:DT} after squaring; the bound is
deterministic, so it survives the expectation. For
Corollary~\ref{cor:Dfinite}, Minkowski's inequality in $L^2$ gives
$(\E\Gamma_t^2)^{1/2}\le\sum_{u\le t}\eta_u(\E\Fnorm{G_u}^2)^{1/2}
\le\Gamma_\infty$. Unrolling the accumulator as in the proof of
Lemma~\ref{lem:Beta} and using $\Fnorm{\tilde E_s}\le(1-\beta_1)\Gamma_s
\le(1-\beta_1)\Gamma_t$ for $s\le t$ gives
$\Fnorm{\tilde B_t}\le(1-\beta_1)\Gamma_t+\gamma\Gamma_t\le2\Gamma_t$, so by
\eqref{eq:Sbound},
$(\E\Fnorm{\Delta_t}^2)^{1/2}\le\big(1+\tfrac{2\kappa}{1-\beta_1}\big)
\Gamma_\infty<\infty$ uniformly in $t$; no schedule regularity is needed. For
$\eta_u=\Theta(1/u)$ and $(\E\Fnorm{G_u}^2)^{1/2}=O(u^{-\epsilon})$ the series
$\sum_u u^{-1-\epsilon}$ converges. \qed

\subsection*{Proof of Proposition~\ref{prop:lower}}

Write $g=(g_1,g_2)$ with $g_1=1/\sqrt{1+c^2}$, $g_2=c/\sqrt{1+c^2}$, and note
$g_2/g_1=c$ and $\Fnorm{g}=1$. Top-$1$ selection acts coordinatewise: if the
first coordinate is selected at step $s$ then
$\tilde M_s=(\tilde b^1_s,0)$ and $\tilde E_s=(0,\tilde b^2_s)$, so
\eqref{eq:recursion-eta} decouples into
\[
  \tilde b^1_t=\beta_1\tilde b^1_{t-1}+(1-\beta_1)\eta_tg_1,
  \qquad
  \tilde b^2_t=\gamma\tilde b^2_{t-1}+(1-\beta_1)\eta_tg_2 ,
\]
whence $\tilde b^1_t=(1-\beta_1)g_1\sum_{i<t}\beta_1^{i}\eta_{t-i}$ and
$\tilde b^2_t=(1-\beta_1)g_2\sum_{j<t}\gamma^{j}\eta_{t-j}$, both positive. We
check by induction that the first coordinate is indeed selected at every step,
i.e.\ $\tilde b^1_t>\tilde b^2_t$. Since $\eta$ is nonincreasing,
$\sum_{i<t}\beta_1^{i}\eta_{t-i}\ge\eta_t(1-\beta_1^{t})/(1-\beta_1)
\ge\eta_t$; since $\eta_{t-j}\le\rho^{j}\eta_t$,
$\sum_{j<t}\gamma^{j}\eta_{t-j}\le\eta_t/(1-\gamma\rho)$. Hence
$\tilde b^1_t/\tilde b^2_t\ge(1-\gamma\rho)g_1/g_2=(1-\gamma\rho)/c>1$ by the
choice of $c$, and the induction closes.

Therefore $S_t=(0,\sum_{s\le t}\tilde b^2_s)$ and, exchanging the order of
summation as in the proof of Theorem~\ref{thm:leak},
$\sum_{s\le t}\tilde b^2_s=(1-\beta_1)g_2\sum_{u\le t}\eta_u
(1-\gamma^{\,t-u+1})/(1-\gamma)$. By \eqref{eq:pi} the second coordinate of
$\Delta_t$ is $[\gamma\tilde b^2_t+(1-\gamma)\sum_{s\le t}\tilde b^2_s]
/(1-\beta_1)$, all terms positive, so
$\Fnorm{\Delta_t}\ge g_2\sum_{u\le t}\eta_u(1-\gamma^{\,t-u+1})$, and
$\sum_{u\le t}\eta_u\gamma^{\,t-u+1}\le\eta_1\gamma/(1-\gamma)$ gives
\eqref{eq:lower}. Finally $g_2=c/\sqrt{1+c^2}<1/\sqrt2$ because $c<1-\gamma\rho\le1$. \qed

\subsection*{Proof of Theorem~\ref{thm:floor}}

Applying \citet[Lemma~3.1]{stich2018} to the virtual sequence of
Proposition~\ref{prop:pi} gives
\[
  a_{t+1}\;\le\;\Bigl(1-\tfrac{\mu\eta_t}{2}\Bigr)a_t+\eta_t^2G^2
  -\eta_te_t+\eta_t(\mu+2L)\,\E\Fnorm{\Delta_t}^2,
\]
with $a_t=\E\Fnorm{\hat X_t-X^\star}^2$ and $e_t=\E f(X_t)-f^\star$. The
substitution is legitimate because that lemma uses the deviation between the
true and virtual iterates only through its squared norm, and
Proposition~\ref{prop:pi} identifies that deviation as $\Delta_t$ exactly. The
recursion \eqref{eq:recursion-eta} is indexed from $t=1$, and $\Delta_0=0$, so
the initial condition is $a_1=\E\Fnorm{X_1-X^\star}^2$.

Under the hypothesis the last term is at most $\eta_t(\mu+2L)D$, which carries
the \emph{same} power of $\eta_t$ as the descent term $-\eta_te_t$, so the two
combine:
\[
  a_{t+1}\;\le\;\Bigl(1-\tfrac{\mu\eta_t}{2}\Bigr)a_t+\eta_t^2G^2
  -\eta_t\bigl(e_t-(\mu+2L)D\bigr).
\]
This is the hypothesis of \citet[Lemma~3.3]{stich2018} with $A=G^2$, $B=0$ and
$e_t$ replaced by $\hat e_t:=e_t-(\mu+2L)D$. Its conclusion gives
\[
  \frac1{W_T}\sum_{t<T}w_t\hat e_t\;\le\;\frac{\mu a^3}{8W_T}a_1
  +\frac{4T(T+2a)}{\mu W_T}G^2 .
\]
Adding $(\mu+2L)D$ back to both sides and using convexity of $f$, so that
$\E f(\bar X_T)-f^\star\le\frac1{W_T}\sum_{t<T}w_te_t$, gives the first claim.

For the second claim, if instead $\E\Fnorm{\Delta_t}^2\le\eta_t^2C$ then the
compression term is $\eta_t(\mu+2L)\eta_t^2C=\eta_t^3(\mu+2L)C$, an $\eta_t^3B$
term with $B=(\mu+2L)C$, which is the form assumed in
\citet[Lemma~3.3]{stich2018}; the conclusion of \citet[Thm.~2.4]{stich2018} then
holds with $\frac{4\alpha}{\alpha-4}\frac{d^2}{k^2}$ replaced by $C$. \qed

\end{document}